\documentclass[pmlr]{jmlr}

\RequirePackage{graphicx}
\usepackage{booktabs}
\usepackage{longtable}

\makeatletter
\def\set@curr@file#1{\def\@curr@file{#1}} 
\makeatother
\usepackage[load-configurations=version-1]{siunitx} 

\usepackage{amsmath}
\usepackage{graphicx}
\usepackage{xspace}
\usepackage{booktabs, multirow}
\usepackage[most]{tcolorbox}
\usepackage{listings}
\usepackage{soul}
\usepackage{threeparttable}
\usepackage{makecell}
\definecolor{JungleGreen}{RGB}{41,171,135}
\newcommand{\mname}{\textit{C-Reason}\xspace}
\renewcommand{\thefootnote}{\fnsymbol{footnote}}
\makeatletter
\renewcommand{\@makefnmark}{\textsuperscript{\normalfont\textcolor{red}{\@thefnmark}}}
\makeatother

\theorembodyfont{\upshape}
\theoremheaderfont{\scshape}
\theorempostheader{:}
\theoremsep{\newline}

\jmlrvolume{340}
\jmlryear{2026}
\jmlrworkshop{Machine Learning for Healthcare}
\jmlrproceedings{PMLR}{Proceedings of Machine Learning Research}

\title[C-Reason]{Enhancing LLMs' Clinical Reasoning with Real-World Data from a Nationwide Sepsis Registry}

\author{%
    \Name{Junu Kim}\textsuperscript{1} \Email{kjune0322@kaist.ac.kr} \AND
    \Name{Chaeeun Shim}\textsuperscript{1} \Email{chaeeun@kaist.ac.kr} \AND
    \Name{Sungjin Park}\textsuperscript{2}$^\star$ \Email{jinpark@microsoft.com} \AND
    \Name{Su Yeon Lee}\textsuperscript{3} \Email{lsy5013@naver.com} \AND
    \Name{Gee Young Suh}\textsuperscript{4} \Email{suhgy@skku.edu} \AND
    \Name{Seong Jin Choi}\textsuperscript{5} \Email{seongjin2300@gmail.com} \AND
    \Name{Song Mi Moon}\textsuperscript{5} \Email{moon7796@hanmail.net} \AND
    \Name{Kyoung-Ho Song}\textsuperscript{5} \Email{khsongmd@gmail.com} \AND
    \Name{Eu Suk Kim}\textsuperscript{5} \Email{eskim@snubh.org} \AND
    \Name{Hong Bin Kim}\textsuperscript{5} \Email{hbkimmd@snu.ac.kr} \AND
    \Name{Sejoong Kim}\textsuperscript{5} \Email{sejoong2@snu.ac.kr} \AND
    \Name{Chami Im}\textsuperscript{5} \Email{chami0921@gmail.com} \AND
    \Name{Dong-Wan Kang}\textsuperscript{5} \Email{dwkang0201@gmail.com} \AND
    \Name{Yong Soo Kim}\textsuperscript{5} \Email{kk35077@gmail.com} \AND
    \Name{Hee-Joon Bae}\textsuperscript{5} \Email{braindoc@snu.ac.kr} \AND
    \Name{Sung Yoon Lim}\textsuperscript{5}* \Email{nucleon727@gmail.com} \AND
    \Name{Han-Gil Jeong}\textsuperscript{5}* \Email{han.g.jeong@gmail.com} \AND
    \Name{Edward Choi}\textsuperscript{1}* \Email{edwardchoi@kaist.ac.kr}
    \AND
    \parbox{\textwidth}{\normalfont\mdseries\upshape\centering\footnotesize
        \textit{--- and on behalf of the Korean Sepsis Alliance (KSA) Investigators ---}\\[0.5ex]
        \textsuperscript{1}\,Korea Advanced Institute of Science and Technology\\
        \textsuperscript{2}\,Microsoft\\
        \textsuperscript{3}\,Asan Medical Center, University of Ulsan College of Medicine\\
        \textsuperscript{4}\,Samsung Medical Center, Sungkyunkwan University School of Medicine\\
        \textsuperscript{5}\,Seoul National University Bundang Hospital, Seoul National University College of Medicine\\
        $^\star$\,Work done at KAIST\\
        *\,Co-corresponding authors}%
}
\begin{document}

\maketitle

\vspace{-3em}
\begin{abstract}
    Although large language models (LLMs) have demonstrated impressive reasoning capabilities across general domains, their effectiveness in real-world clinical practice remains limited. This limitation is likely due to insufficient exposure to real-world clinical data during training, as such data are typically excluded because of privacy concerns. To address this gap, we trained an LLM on real-world clinical data from a nationwide sepsis registry and evaluated the reasoning improvements across diverse datasets and tasks. The trained model demonstrated strong clinical reasoning performance on in-domain test sets, supported by both quantitative metrics and expert evaluations. Moreover, these enhanced reasoning capabilities generalized to an external sepsis dataset involving different tasks and patient cohorts, an open-ended antibiotic consultation task, and a disease beyond sepsis. Future research should focus on training LLMs on large-scale, multi-disease clinical datasets to enable more powerful and general-purpose clinical reasoning models.
\end{abstract}

\section{Introduction}

Unlike traditional machine learning models, large language models (LLMs) can generate various forms of reasoning in natural language.
This capability enables them to imitate the clinical reasoning processes of medical experts, offering several advantages.
First, by revealing the rationale behind their decisions, LLMs help experts better understand and trust the decisions.
Second, the reasoning capabilities enable strong performance across a range of tasks, including medical licensing exams \citep{nori2023can, singhal2025toward} and diagnostic applications \citep{savage2024diagnostic, wang2025medical, cabral2024clinical}.
However, LLMs still exhibit limited clinical reasoning capabilities in tasks that reflect real-world clinical practice, such as those involving rare conditions, adherence to clinical guidelines, and interpretation of structured patient data \citep{kim2025limitations, hager2024evaluation, reese2024limitations}.

One possible explanation for this limited clinical reasoning capability is LLMs' insufficient exposure to real-world clinical data during training.
While medical experts rely on a combination of medical knowledge and accumulated clinical experience to perform clinical reasoning, LLMs are typically trained on web-based corpora, including textbooks and journal articles rich in medical knowledge \citep{touvron2023llama, brown2020language}.
However, due to privacy restrictions and limited data-sharing practice, real-world clinical data that embody clinical experience is rarely available online.
Given that LLM performance in a domain depends on the amount of related training data \citep{kandpal2023large}, this insufficient exposure may hinder their ability to reason effectively in real-world clinical settings.

Thus, we trained an LLM to enhance its clinical reasoning capabilities by leveraging real-world clinical data and evaluated whether these capabilities improved across multiple evaluation settings.
We applied reinforcement learning, which has been widely used to train reasoning capabilities in general-domain LLMs \citep{guo2025deepseek, shao2024deepseekmath, wang2023math, kumar2024training}, to a nationwide multicenter sepsis registry \citep{jeon2019characteristics} (Figure~\ref{fig:method}).
Using this approach, we trained Phi-4 LLM \citep{abdin2024phi}, resulting in \mname (\textbf{C}linical\textbf{-Reason}er).
Enhanced clinical reasoning capabilities were observed on the in-domain test set of the sepsis registry, as confirmed through both quantitative evaluation and expert assessment.

Notably, the trained model demonstrates improved clinical reasoning not only within the sepsis registry but also across a range of tasks and datasets.
First, when evaluated on a separate sepsis dataset involving a different cohort and tasks than those used during training, \mname showed improved clinical reasoning.
Second, in an open-ended clinical reasoning evaluation involving consultations on antibiotics use for patients with infections, experts consistently preferred the responses generated by \mname over those produced by Phi-4.
Third, to assess the model's reasoning capabilities beyond sepsis, we conducted experiments on two additional cohorts: hospitalized patients with a feature set related to acute kidney injury, and patients from a nationwide, multicenter stroke registry.
Performance improvements were only observed on the hospitalized cohort, indicating partial cross-disease generalizability.
Overall, these results suggest that training an LLM on real-world clinical data can enhance its clinical reasoning across multiple settings.
This finding highlights the importance of future work focusing on large-scale, multi-disease training to develop more robust and general-purpose clinical reasoning LLMs.

\subsection*{Generalizable Insights about Machine Learning in the Context of Healthcare}

\begin{itemize}
    \item \textbf{Real-world clinical records can directly train reasoning, without external teacher rationales.} A simple reinforcement learning setup over masked-feature questions from a nationwide registry improves an LLM's performance on real-world clinical tasks designed by clinicians. Because no proprietary model generates the training rationales, the pipeline is reproducible and survives data-governance review\footnote{\url{https://github.com/starmpcc/c-reason}}.
    \item \textbf{Improvements in reasoning traces, not just answers, transfer across cohorts, tasks, and modalities.} Although the model is trained only on a nationwide sepsis registry, clinician-designed evaluations on a separate sepsis cohort (MIMIC-III), an open-ended antibiotic-consultation task, and an AKI cohort all show gains over the base model. This argues that gains reflect reasoning patterns learned from real-world clinical data, not memorized dataset correlations.
    \item \textbf{Clinician preference validates that gains are in reasoning quality, not just answer accuracy.} Blinded preference evaluations conducted by eight clinicians consistently favor the trained model's reasoning traces over those of the base model. This shows that the model is not merely guessing the right answer; it is reasoning in a way that clinicians find more clinically sound.
\end{itemize}

\section{Related Works}

\textbf{Reasoning Capability of LLMs in General Domain}

Recently, reasoning-oriented LLMs, such as OpenAI's o3-mini-high \citep{o3-mini} and Deepseek-R1 \citep{guo2025deepseek}, have demonstrated impressive performance in domains like mathematical olympiads, graduate-level science problems, and competitive programming tasks.
LLMs develop this capability through pretraining on large-scale web corpora to build language proficiency and general world knowledge, followed by reinforcement learning focused on reasoning using questions in mathematics, science and programming \citep{kumar2024training, wang2023math, shao2024deepseekmath, guo2025deepseek}.
In this second phase, the model generates one or more reasonings for each problem, which are then evaluated by an independent model \citep{wang2023math, kumar2024training} or a scoring criterion \citep{guo2025deepseek, shao2024deepseekmath} to assign a reward.
The model is subsequently fine-tuned via reinforcement learning algorithms to generate higher-reward reasoning \citep{schulman2017proximal, shao2024deepseekmath}.

\textbf{Clinical Reasoning Capability of LLMs}

Reasoning tasks in general domains such as mathematics, science, or programming are typically well-defined, have explicit solutions, and rely on deterministic logic.
On the other hand, clinical reasoning is inherently patient-specific, implicit, and often heuristic \citep{durning2011context, thinking2008making, hicks2011heuristic}.
These characteristics make clinical reasoning particularly challenging for LLMs, as they are typically trained on a limited amount of clinical data.
Consequently, several studies have highlighted the limitations of LLMs in performing tasks that reflect real-world clinical practice \citep{kim2025limitations, hager2024evaluation, reese2024limitations}.
Recently, several attempts have been made to enhance the clinical reasoning capabilities of LLMs using real-world clinical data \citep{kwon2024large, wu2024instruction}.
However, these methods typically involve generating questions from clinical data and then training the model using the reasoning provided by a powerful external model (e.g., GPT-5 \citep{singh2025openai}).
This reliance on an external model complicates the overall model development process, for example by requiring approval from a Data Review Board.
In contrast, our work focuses on training an LLM to improve its clinical reasoning solely using real-world clinical data, without relying on external models.

\begin{figure*}[!htbp]
    \includegraphics[width=1.0\linewidth,trim=0 255 0 130, clip]{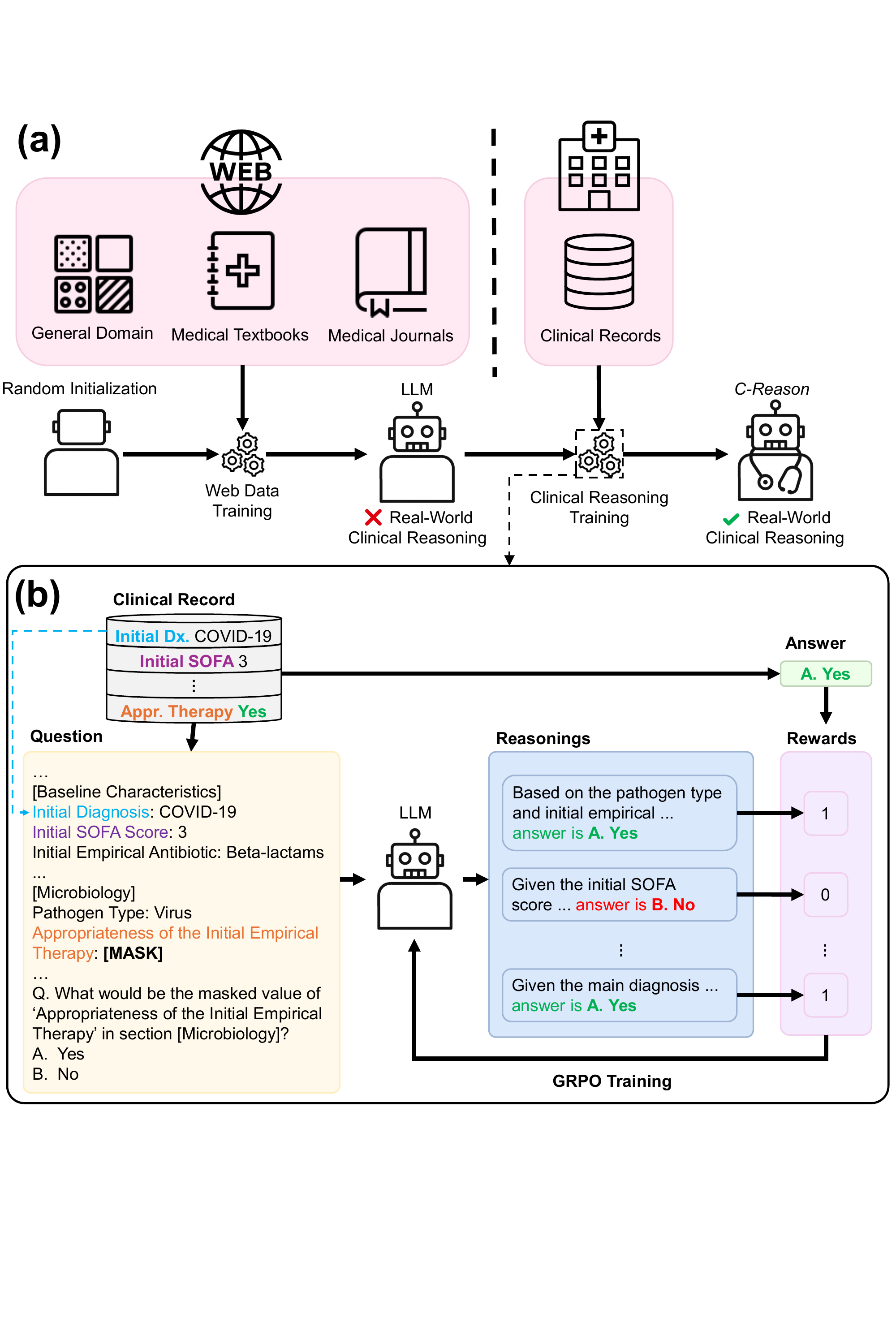}
    \caption{(a) Motivation and Approach. LLMs are primarily trained on web corpora, which leads to insufficient exposure to real-world clinical data and results in limited clinical reasoning capabilities. To address this gap, we further trained an LLM on clinical data, thereby enhancing its real-world clinical reasoning performance.
        (b) Illustration of the Proposed Method. First, multiple-choice denoising questions are generated from the clinical data (sepsis registry). Then, the LLM generates multiple reasonings for each question and the rewards are calculated based on their correctness. Finally, the model is optimized using the GRPO algorithm \citep{shao2024deepseekmath}.
    }
    \label{fig:method}
\end{figure*}

\section{Methods}
Clinical data consist of multiple feature–value pairs (e.g., Initial SOFA - 3) for each patient.
Training LLMs for reasoning usually involves prompting the model with reasoning-intensive questions \citep{guo2025deepseek, shao2024deepseekmath, wang2023math, kumar2024training}.
Therefore, generating such questions from clinical data is essential to enhance clinical reasoning of LLMs.
One possible approach is to construct open-ended questions.
However, open-ended questions make it difficult to establish consistent reward criteria, which can result in unstable training \citep{skalse2022defining}.
To address this, Deepseek-R1 \citep{guo2025deepseek} limited its training data to short-answer questions with well-defined answers, such as those involving math, coding, and logical reasoning.
In this setup, rewards were given only if the model's reasoning led to the correct answer, enabling stable training.
Building on this approach, we construct multiple-choice denoising questions by masking the value of a single feature in each patient's data.
Then the model is prompted to infer the masked value based on the remaining visible feature-value pairs.
While this task resembles standard imputation in simple time-series clinical data, it requires more sophisticated clinical reasoning when applied to complex datasets such as clinical registries.
As illustrated in Figure \ref{fig:method}-(b), inferring the masked value, such as the initial empirical therapy, requires the model to consider inter-feature relationships and dependencies among variables like the initial diagnosis or the selection of empirical antibiotics.
This process encourages the model to learn deeper clinical reasoning capabilities.
We emphasize that this denoising task is used as a training vehicle for eliciting clinical reasoning, chosen because it admits a verifiable reward, and is not itself intended as a clinically meaningful task.
The clinical claims of this work are therefore evaluated on downstream clinician-designed tasks that reflect real-world clinical practice, not on the denoising objective.
We regard masked-feature denoising as an easier, lower-level form of clinical reasoning that infers inter-feature dependencies from partial patient data, and the downstream clinician-designed tasks as a harder form that applies the same reasoning to higher-level judgments.
Improving on the harder tasks by training on the easier one is consistent with easy-to-hard generalization, a well-documented phenomenon in reasoning-oriented reinforcement learning where models trained on more tractable, verifiable problems reliably improve on harder, unseen ones \citep{sun2024easy, hase2024unreasonable}.
We adopt reinforcement learning rather than supervised fine-tuning because the latter would require reasoning traces obtained either from medical experts (costly and time-consuming) or from a powerful external model (privacy-constrained), whereas reinforcement learning only requires a verifiable correctness signal that our denoising questions already provide.
For each question, the model generates multiple reasonings and only the ones that led to the correct answer receive a reward.
Using the Group Relative Policy Optimization (GRPO) \citep{shao2024deepseekmath}, a reinforcement learning algorithm used to train Deepseek-R1 \citep{guo2025deepseek}, the model is optimized to generate reasonings that lead to high rewards.
An overview of the process is illustrated in Figure \ref{fig:method}-(b).

Clinical data often contain substantial redundancy, where multiple features may encode overlapping information.
As a result, some denoising targets can be inferred directly from closely related features without requiring genuine clinical reasoning.
To prevent the model from exploiting such surface-level correlations rather than developing clinical reasoning, we pre-compute mutual information between all pairs of features and exclude as masking targets any feature whose context retains a highly correlated feature (mutual information $>$ 0.5).
For example, if hospital length of stay is masked while both admission and discharge dates remain visible, the answer can be inferred without clinical reasoning; such configurations are filtered out.
In addition, certain masking configurations could leak future information from the patient's trajectory, allowing the model to answer via temporal shortcuts rather than clinical reasoning; we apply additional filtering to remove such cases.
Further details of these filtering procedures are provided in Appendix \ref{apd:method}.

Based on the methodology described above, we generated questions from a sepsis registry maintained by the Korean Sepsis Alliance (KSA), covering patients who were enrolled between September 2019 and December 2021 \citep{jeon2019characteristics}.
This nationwide dataset includes adult sepsis patients (aged 18 years or older) from 16 tertiary or university-affiliated hospitals across South Korea.
The dataset comprises information of 11,981 patients and 691 features, such as demographics, laboratory results, treatments, and outcomes.
A random subset of 1,000 patients was held out to form a test set.
From the remaining data, 30,000 multiple-choice questions were constructed to train Phi-4 \citep{abdin2024phi}\footnote{At the time of our experiments, Phi-4 was the most capable general-purpose LLM we were able to train on our 8×A100 GPU setup.}, resulting in the model \mname.
Our pipeline has no model-specific components; to confirm that the gains are not specific to Phi-4, we additionally apply it to another base model, Llama-3.1-8B-Instruct \citep{grattafiori2024llama} in Appendix~\ref{apd:base}.
Data statistics, sample questions, and further implementation details are provided in the supplementary material (Appendices \ref{apd:stat}, \ref{apd:samples}, and \ref{apd:method}).
The study was approved by the Institutional Review Boards of the corresponding institution.


\begin{table*}[!t]\centering
\caption{Performance Evaluation Results. We report accuracy for multiple-choice tasks, and both accuracy and F1 score (in parentheses) for binary prediction tasks. For each task, the highest score is bolded, and the second-highest is underlined. 
Results for each dataset are discussed in the following sections: \textcolor{red}{Red} (Section~\ref{3.1}), \textcolor{blue}{Blue} (Section~\ref{3.2}), and \textcolor{JungleGreen}{Green} (Section~\ref{3.4}).
}\label{tab:main_perf}
\setlength{\tabcolsep}{0.2em}
\begin{threeparttable}
\resizebox{\textwidth}{!}{
\begin{tabular}{lrrrrr}\toprule[3pt]
Dataset &\color{red}{Sepsis Registry} &\color{blue}{MIMIC-III} &\color{JungleGreen}{Hospitalized Cohort} &\color{JungleGreen}{Stroke Registry} \\\cmidrule{1-5}
Task &Den.\tnote{1} (Avg.) &Measurement Pred.\tnote{2} (Avg.) &Den. (Avg.) &Den. (Avg.) \\\cmidrule{1-5}
Phi-4 &\ul{0.712} &\ul{0.623} &\ul{0.654} &\ul{0.739} \\\cmidrule{1-5}
\mname &\textbf{0.864} &\textbf{0.747} &\textbf{0.796} &\textbf{0.833} \\\cmidrule[3pt]{1-5}
Dataset &\multicolumn{4}{c}{\color{red}{Sepsis Registry}} \\\cmidrule{1-5}
Task &Initial Lactate Den. &ECOG at Discharge Den. &Discharge Status Den. &App. Ini. Emp.\tnote{3} \\\cmidrule{1-5}
Phi-4 &0.335 &0.379 &0.790 &0.693 \\\cmidrule{1-5}
\mname &\textbf{0.801} &\textbf{0.707} &\ul{0.849} &\textbf{0.896} \\\cmidrule{1-5}
o3-mini-high &\ul{0.787} &0.560 &\textbf{0.879} &\ul{0.833} \\\cmidrule{1-5}
Deepseek-R1 &0.680 &\ul{0.661} &0.760 &0.688 \\\cmidrule{1-5}
QwQ-32B &0.767 &0.605 &0.767 &0.697 \\\cmidrule{1-5}
Qwen2.5-14B-Instruct &0.330 &0.426 &0.831 &0.667 \\\cmidrule{1-5}
DeepSeek-R1-Distill-Qwen-14B &0.498 &0.501 &0.845 &0.761 \\\cmidrule{1-5}
Meditron3-Phi4-14B &0.416 &0.397 &0.756 &0.708 \\\cmidrule[3pt]{1-5}
Dataset &\color{blue}{MIMIC-III} &\color{JungleGreen}{Hospitalized Cohort} &\multicolumn{2}{c}{\color{JungleGreen}{Stroke Registry}} \\\cmidrule{1-5} 
Task &In-Hospital Mortality Pred. &48h AKI Pred. &3-months mRS Pred. &1-year MACE Pred. \\\cmidrule{1-5}
Phi-4 &0.627 (0.231) &0.640 (0.328) &0.525 &0.330 (0.210) \\\cmidrule{1-5}
\mname &\textbf{0.862 (0.274)} &\textbf{0.933 (0.599)} &0.635 &\textbf{0.708 (0.198)} \\\cmidrule{1-5}
o3-mini-high &\ul{0.732 (0.264)} &- &- &- \\\cmidrule{1-5}
Deepseek-R1 &0.360 (0.200) &- &- &- \\\cmidrule{1-5}
QwQ-32B &0.162 (0.183) &0.842 (0.500) &\textbf{0.656} &0.244 (0.197) \\\cmidrule{1-5}
Qwen2.5-14B-Instruct &0.424 (0.202) &\ul{0.885 (0.585)} &0.553 &\ul{0.442 (0.218)} \\\cmidrule{1-5}
DeepSeek-R1-Distill-Qwen-14B &0.467 (0.203) &0.738 (0.407) &\ul{0.649} &0.283 (0.206) \\\cmidrule{1-5}
Meditron3-Phi4-14B &0.596 (0.252) &0.702 (0.349) &0.426 &0.467 (0.198) \\\midrule
\bottomrule
\end{tabular}
}
\begin{tablenotes}
\footnotesize
  \item[1] Denoising
  \item[2] Prediction
  \item[3] Appropriateness of Initial Empirical Therapy
\end{tablenotes}
\end{threeparttable}

\end{table*}

\section{Results}

\begin{figure}[!t]
    \includegraphics[width=1.0\linewidth]{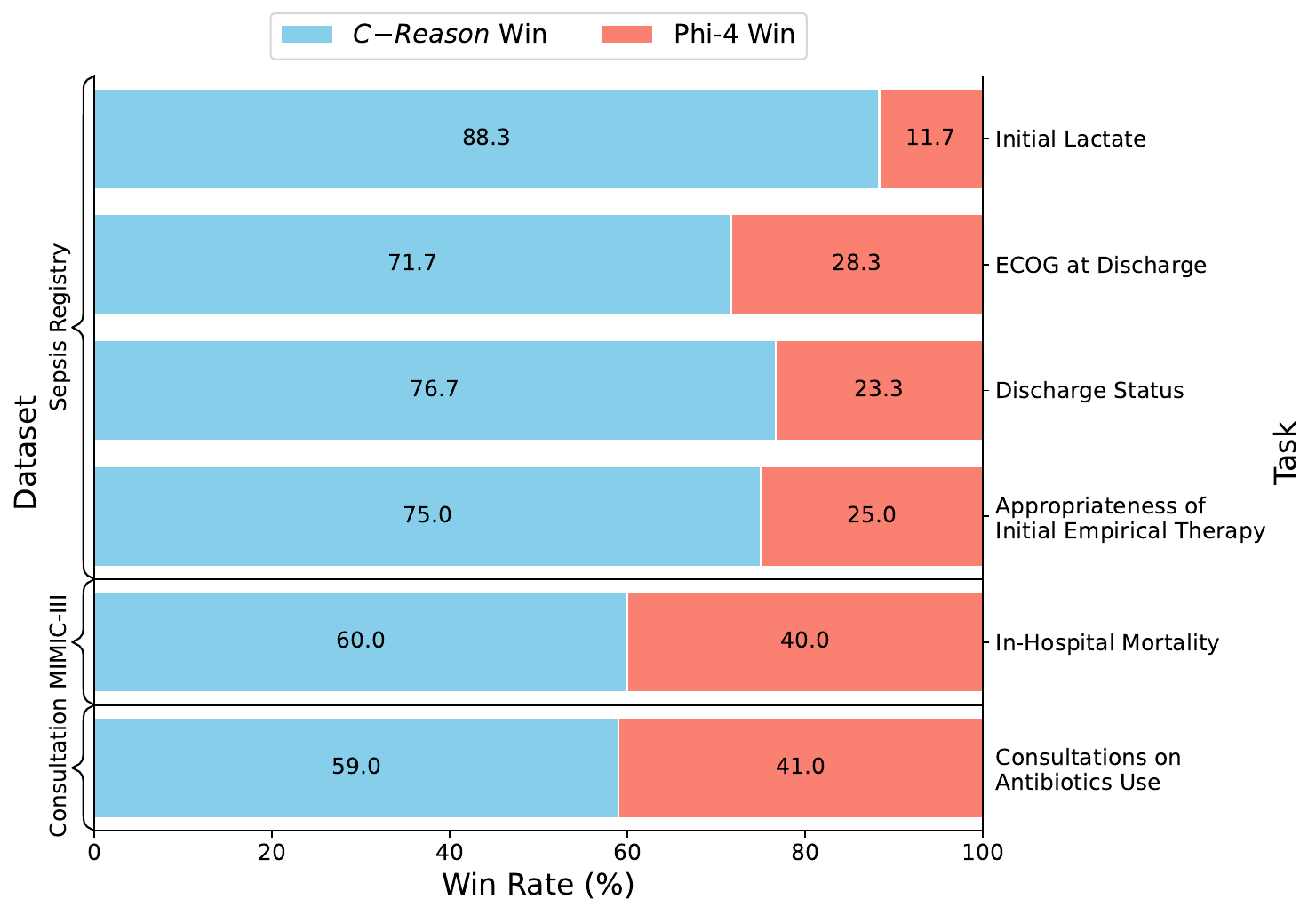}
    \caption{Reasoning Expert Evaluation Results. We report win rate (\%) for each task.}
    \label{fig:human_eval}
\end{figure}

A central concern for verifiable-reward reinforcement learning is that a model may arrive at the correct final answer through flawed reasoning, in which case accuracy gains do not necessarily reflect improved reasoning.
Because our primary claim is that training on real-world clinical data improves clinical reasoning, we treat blinded clinician preference over the generated reasoning traces (Figure \ref{fig:human_eval}) as our principal line of evidence.
Accuracy on multiple-choice tasks (Table \ref{tab:main_perf}) serves as supporting evidence; paired McNemar's tests confirm that \mname's gains over Phi-4 are statistically significant ($p<0.001$) across all reported tasks in the table.

\subsection{Clinical Data Training Improves Clinical Reasoning on In-Domain Data}\label{3.1}
To assess whether our approach improved the LLM's clinical reasoning capabilities, we examined \mname on an in-domain test set from the sepsis registry.
We compared \mname against seven baseline models, including its base model (Phi-4 \citep{abdin2024phi}), state-of-the-art general domain reasoning models (o3-mini-high \citep{o3-mini}, DeepSeek-R1 \citep{guo2025deepseek}, QwQ-32B \citep{qwq}), and models of comparable sizes (Qwen2.5-14B-Instruct \citep{yang2024qwen2}, DeepSeek-R1-Distill-Qwen-14B \citep{guo2025deepseek}, Meditron3-Phi4-14B \citep{chen2023meditron}).
Due to computational resource constraints, we report denoising performance across all available features only for Phi-4 and \mname.
For the remaining models, evaluation was limited to the four features deemed most important by clinicians.
The results are displayed on Table \ref{tab:main_perf}, and detailed per-feature results for Phi-4 and \mname are provided in the supplementary material (Appendix \ref{apd:res}).

In terms of performance, \mname significantly outperformed its base model, Phi-4.
Furthermore, the model consistently outperformed all other models of comparable size and matched or exceeded state-of-the-art models in general-domain reasoning, demonstrating the effectiveness of training on real-world clinical data.

In clinical reasoning, while accurate decision-making is essential, the underlying rationale is equally important, as it helps experts understand and trust the model's outputs \citep{holzinger2017we}.
Therefore, we conducted an expert evaluation involving three intensivists with clinical expertise in sepsis and critical care.
Each expert was presented with 20 cases per task across the four selected tasks, which were identical for all evaluators.
To minimize cognitive load, evaluators were asked to perform a binary preference task, in which they compared the clinical reasoning of Phi-4 and \mname side by side.
The results of this evaluation are presented in Figure \ref{fig:human_eval}.

\begin{figure*}[!htbp]
    \centering
    \begin{tcolorbox}
        \footnotesize
        \textbf{Model Input}\\
        \input{case/initemp_input}

        \noindent\hrulefill

        \textbf{Phi-4's Response}\\
        \input{case/initemp_phi4}

        \noindent\hrulefill

        \textbf{\mname's Response}\\
        \input{case/initemp_grpo}
    \end{tcolorbox}
    \caption{Case Analysis - Appropriateness of Initial Empirical Therapy}
    \label{fig:initemp}

\end{figure*}

Overall, intensivists showed a significant preference for \mname's responses ($p<0.0001$).
Notably, they reported a substantial logical improvement in the Appropriateness of Initial Empirical Therapy task, which judges the appropriateness of the initial antibiotic selection.
In the case shown in Figure \ref{fig:initemp}, the patient was diagnosed with COVID-19, and initial antibiotic selection was beta-lactams.
Phi-4 evaluated this therapy as inappropriate, citing that SARS-CoV-2 is a virus and that beta-lactams lack efficacy against viral infections.
In contrast, \mname considered the therapy appropriate, noting that empirical antibiotics are commonly administered in patients with suspected sepsis, not only before the underlying etiology is confirmed but also after a confirmed diagnosis of COVID-19, to account for the risk of concurrent bacterial infections \citep{alhazzani2020surviving}.
Collectively, these findings indicate the efficacy of training on real-world clinical data in in-domain scenarios.

\begin{figure*}[!htbp]
    \centering
    \begin{tcolorbox}[before skip=0pt, after skip=0pt]
        \footnotesize
        \textbf{Model Input}\\
        \input{case/consulting_input}

        \noindent\hrulefill

        \textbf{Phi-4's Response}
        \input{case/consulting_phi4}

        \noindent\hrulefill

        \textbf{\mname's Response}
        \input{case/consulting_grpo}
    \end{tcolorbox}
    \caption{Case Analysis - Consultations on Antibiotics Use}
    \label{fig:consulting}
\end{figure*}

\subsection{Trained Clinical Reasoning Generalizes Across Cohorts and Tasks}\label{3.2}

To evaluate \mname's clinical reasoning on a sepsis dataset that differs from its training data in terms of both cohort and task, we used the MIMIC-III database \citep{johnson2016mimic}.
This cohort differs substantially from the training data in terms of geographic region (United States vs. South Korea) and study setting (single-center vs. nationwide multicenter).
Following the approach of a previous study \citep{komorowski2018artificial}, we selected sepsis patients and constructed a time-series feature set consisting of lab values, vital signs, and other measurements sampled at a 4-hour interval\footnote{While MIMIC-IV \citep{johnson2023mimic} is more recent, filtering sepsis cohorts and preprocessing the associated features is highly time- and cost-intensive. Therefore, we adopted the existing filtering and preprocessing pipeline of \citet{komorowski2018artificial}, which was built on MIMIC-III \citep{johnson2016mimic}.}.
Because MIMIC-III is publicly available and widely used in prior work, we cannot rule out that the base models were exposed to portions of it during pretraining, which could in principle confound our out-of-distribution claims.
We note, however, that any such exposure applies equally to Phi-4 and \mname, so it does not bias the relative comparison between them.
Moreover, the specific inputs used in this evaluation are unlikely to have appeared in any pretraining corpus: the preprocessing pipeline of \citet{komorowski2018artificial} is publicly released as code that converts raw MIMIC-III into numeric tables, but the resulting preprocessed tables themselves are not distributed online.
We then implemented a separate step that serializes these numeric tables into textual patient trajectories suitable for LLM input, which was newly written for this work and is not public.
During this serialization, identifiers such as \texttt{hadm\_id} and \texttt{icustay\_id} were explicitly excluded.
Therefore, any residual contamination from MIMIC-III itself is unlikely to materially affect our conclusions.

We sampled a total of 1,000 patients from this dataset.
While the training dataset focused on denoising tasks, here we formulated two prediction tasks instead.
One task involves predicting individual feature values from time-series data up to 24 hours prior, and the other focuses on in-hospital mortality, which is a key clinical outcome.
These differences offer a robust testbed for evaluating whether the  improved clinical reasoning generalizes across both cohort and task.
As in previous experiments, only the performance of Phi-4 and \mname is reported for the feature value prediction task.
Dataset statistics and representative examples are provided in the supplementary material (Appendices \ref{apd:stat} and \ref{apd:samples}), and the experimental results are shown in Table \ref{tab:main_perf}.

The trained model significantly outperformed Phi-4 in the value prediction tasks and surpassed all baselines in the in-hospital mortality task.
These results suggest that the model trained on the sepsis registry can generalize its clinical reasoning to other sepsis datasets.
In addition, we conducted an expert evaluation for the in-hospital mortality prediction task.
The evaluation followed the same protocol as the previous experiment, and the results are presented in Figure \ref{fig:human_eval}.
Responses generated by \mname were preferred over those from Phi-4 by the intensivists, with a win rate of 60\%.
Although the difference did not reach statistical significance ($p = 0.07$), these results provide potential evidence for the generalizability of the trained reasoning capability across both cohorts and tasks.

\subsection{Trained Clinical Reasoning Generalizes to Open-Ended Task}\label{3.3}

Previous evaluations using the sepsis registry and MIMIC-III have primarily focused on multiple-choice question answering tasks.
However, real-world clinical practice often demands open-ended reasoning without predefined answer choices.
To address this, we compared \mname and Phi-4 on an open-ended generative task: consultations on antibiotics use for patients with infection, a task relevant to sepsis.
This task involves generating expert recommendations on the appropriate use of antibiotics, including drug selection, dosing, and duration, in order to minimize resistance and adverse effects.
Each consultation pair consists of a request and a response, where the response includes a summary of patient information and a set of clinical conclusions.
Note that the consultations are unstructured, whereas the training data is in a structured format.
For this task, the models were given the full consultation request and the patient information section of the response, and were asked to generate the conclusions.
We curated 100 consultation pairs from Seoul National University Bundang Hospital, South Korea, collected between January 2023 and January 2025.
The evaluation was performed by four infectious disease specialists and one intensivist.
Each evaluator reviewed 20 non-overlapping consultations and assessed the responses of Phi-4 and \mname using a binary preference format.
The evaluation results are shown in Figure \ref{fig:human_eval}.

The responses of \mname were preferred than those of Phi-4 ($p<0.05$).
In the case analysis shown in Figure \ref{fig:consulting}, a patient was admitted with Influenza A and acute respiratory distress syndrome, and was treated with cefepime for two weeks due to suspected bacterial pneumonia.
Both models appropriately recommended discontinuing cefepime because there was no clear evidence of bacterial infection.
However, as \textit{Candida} was isolated from sputum and urine cultures, Phi-4 recommended immediate antifungal treatment, whereas \mname emphasized reassessment, noting it could represent colonization rather than a true infection.
The \mname's response aligns more closely with antibiotic stewardship principles.
These results suggest that the trained clinical reasoning generalizes to open-ended generative scenarios involving unstructured data.

\subsection{Trained Clinical Reasoning Generalizes Across Diseases}\label{3.4}

The previous evaluations primarily focused on sepsis or infection, which closely aligned with the sepsis registry used for training.
To assess whether the trained clinical reasoning generalizes beyond sepsis, we conducted evaluations using two additional cohorts: a hospitalized cohort with a feature set related to acute kidney injury (AKI), and a stroke cohort.
The hospitalized cohort consists of all adult inpatients from two tertiary hospitals in South Korea between 2013 and 2017 who had serum creatinine measurements available for more than two days during their hospital stay \citep{im2024case}.
The stroke cohort was derived from the Clinical Research Collaboration for Stroke in Korea (CRCS-K), a nationwide multicenter registry that has been collecting data since April 2008 \citep{kim2015case}.
We sampled 1,000 patients from each dataset and performed the feature denoising tasks using the same method as with the sepsis registry.
Additionally, for the hospitalized cohort, we performed 48-hour AKI prediction, while for the stroke cohort, we conducted 3-month modified Rankin Scale (mRS) \citep{farrell1991united} prediction and 1-year Major Adverse Cardiovascular Events (MACE) prediction.
As in previous evaluations, we report the denoising performance only for Phi-4 and \mname.
Due to data usage restrictions, these datasets could not be transferred outside the hospital environment.
Therefore, we were unable to evaluate o3-mini-high (proprietary) and Deepseek-R1 (due to its large model size) on the prediction tasks.
Data statistics and representative examples are provided in the supplementary material (Appendices \ref{apd:stat} and \ref{apd:samples}).
The results are shown in Figure \ref{tab:main_perf}.

As a result, \mname outperformed Phi-4 and all baseline models on the denoising task and the AKI prediction task.
In the 3-month mRS prediction task, its performance was comparable to that of the best-performing baseline and superior to Phi-4.
Although accuracy improved significantly in the 1-year MACE prediction task, the F1 score declined, indicating that the model failed to generalize to this task.
We hypothesize that this may be due to differences in disease similarity: sepsis and AKI exhibit high comorbidity and share several clinical characteristics \citep{aguilar2024sepsis}, whereas the overlap between sepsis and stroke is relatively limited \citep{shao2019risk}.
Overall, these results provide empirical evidence suggesting partial generalizability across different diseases.
We explicitly note the 1-year MACE result as a negative finding that bounds the scope of our claim: the trained reasoning does not generalize uniformly to diseases with limited clinical overlap with sepsis.

\section{Discussion}

Recent large language models (LLMs) have achieved remarkable performance on general-domain reasoning tasks. However, their clinical reasoning capabilities in real-world practice remain limited \citep{kim2025limitations, hager2024evaluation, reese2024limitations}. These limitations may stem from insufficient exposure to real clinical data during training. To address this gap, we trained an LLM on real-world clinical data to enhance its clinical reasoning abilities and evaluated it across diverse settings. Compared with Phi-4, the pretraining baseline, \mname demonstrated significant improvements in both quantitative metrics and expert assessments across multiple scenarios, including in-domain data, a distinct sepsis dataset, an open-ended task involving unstructured data, and a task involving different disease.

Medicine is inherently complex and highly interconnected. Patients often present with multiple coexisting conditions that interact in unpredictable ways, requiring clinicians to integrate diverse information across organ systems and disease categories. This complexity underscores the limitations of developing isolated models tailored to individual conditions. In light of this, the need for a general-purpose clinical reasoning LLM becomes increasingly apparent. Given that \mname achieved meaningful generalizability despite being trained on a single registry, training on multiple datasets spanning diverse diseases would likely yield stronger and more comprehensive clinical reasoning models.

The true potential of these strong and comprehensive clinical reasoning LLMs lies not in static prediction, but in their ability to engage in dynamic, context-aware interaction with clinicians.
Unlike conventional decision support tools, these models may serve as interactive reasoning partners capable of exploring alternative hypotheses, clarifying clinical thought processes, and providing guideline-based justifications in real time.
In doing so, they have the capacity to augment, rather than replace, expert medical judgment.
We believe that as experts recognize the model's suggestions are grounded in patterns observed in actual patient care, its integration into real-world practice as a credible and supportive tool for nuanced clinical reasoning will become increasingly feasible.

Despite these promising advancements, a significant challenge remains: access to diverse, high-quality clinical data necessary for training such models.
While regulatory constraints such as HIPAA and GDPR limit the sharing of electronic health records across institutions, many additional datasets including registries, proprietary databases, and unpublished research data, also remain inaccessible due to privacy, legal, or institutional barriers.
Building truly generalizable models requires not only expanding access to currently nonpublic datasets, but also developing methods that enable their secure and privacy-preserving use for model training.
Future work should explore privacy-preserving strategies such as federated learning and secure multi-party computation to enable collaborative training without exposing raw patient data.

\paragraph{Limitations}
Our work has several limitations that bound its claims.
First, the training objective is a denoising task, which is not itself a real-world clinical activity; we treat it as a surrogate for eliciting reasoning and evaluate clinical claims on downstream clinician-designed tasks rather than on denoising itself.
Second, cross-disease generalization is partial: gains transferred to sepsis-adjacent settings (infection consultation, AKI) but not uniformly to stroke outcomes, indicating that training on a single registry is insufficient for disease-agnostic clinical reasoning.
Third, our experiments were bounded by computational resources (an 8$\times$A100 setup), so we could not train much larger frontier models; whether the observed gains persist or diminish at that scale remains an open question.
Finally, we did not observe fabricated results in the expert evaluation, but the model occasionally misinterpreted causal relationship (Appendix~\ref{apd:failure}).

\acks{This work was supported by the SNUBH-KAIST Joint Graduate Research Project on AI, Korea Centers for Disease Control and Prevention (grant numbers 2019E280500, \\2020E280700, and 2021-10-026), and Korean Sepsis Alliance (KSA) affiliated with Korean Society of Critical Care Medicine (KSCCM), the Institute of Information \& Communications Technology Planning \& Evaluation (IITP) grant (No.RS-2019-II190075), the InnoCORE Program (N10250156), and National Research Foundation of Korea (NRF) grant (NRF2020H1D3A2A03100945), funded by the Korea government (MSIT).}

\bibliography{chil-sample}

\newpage
\appendix

\section{Implementation Details}\label{apd:method}

\subsection{Multiple-Choice Question Generation}

Clinical data often present challenges such as missing data and redundancy.
Since we use LLMs, strict input formatting is not required, and missing features can simply be omitted. As a result, we did not perform any imputation for missing values.
Redundancy, however, poses a different challenge. When highly relevant information that is strongly correlated with the masked feature is present, the model may exploit surface-level patterns rather than developing meaningful clinical reasoning.
For instance, if the masked feature is the hospital length of stay and both the admission and discharge dates are provided, the model could infer the answer directly instead of reasoning through the clinical context.
To mitigate this issue, we computed the mutual information between all pairs of features and removed those that were highly correlated with the masked target (mutual information $>$ 0.5).
The goal of this filtering is to remove cases where the masked target can be trivially recovered from another feature.
For a binary target, this dependence is immediate: masking ``Discharge within 7 days'' while ``Length of Stay'' is present makes the former fully inferable from the latter, so the mutual information reaches its maximum and the case is filtered.
For a partially deterministic mapping, mutual information captures the partial dependence; for example, when the target is ``Discharge Place'' and an ``Alive or Dead'' feature exists, a deceased patient maps deterministically to the ``Dead'' category while survivors still require inference.
For a pair of (possibly multi-class) features $X$ and $Y$, the mutual information is computed as
\begin{equation*}
    I(X;Y) = \sum_{i}\sum_{j} p(x_i, y_j)\log\frac{p(x_i, y_j)}{p(x_i)\,p(y_j)}.
\end{equation*}
To set the cutoff, clinicians inspected several threshold values between 0.1 and 0.9 on representative tasks and selected 0.5 as an appropriate level.
As a post-hoc check on whether shortcuts were removed, we trained an XGBoost classifier on the four manually designed sepsis registry tasks.
Before mutual-information filtering, some tasks were perfectly solvable (accuracy up to 1.0), indicating complete shortcuts; after filtering, accuracy dropped on all tasks, confirming that the shortcuts were removed.
On average, 2 to 3 direct shortcuts were masked per task.

To generate each question, we converted feature–value pairs into natural language using the format \textit{feature name (unit): value} (e.g., Lactate (mmol/L): 3.1).
We then masked one value (e.g., Lactate (mmol/L): [MASK]) and appended a prompt asking for the original value, along with a set of answer choices.
To encourage meaningful clinical reasoning, we designed the answer choices based on each feature's distribution, aiming for an appropriate level of difficulty.
If the choices are too easy or too difficult, the model may struggle to develop effective reasoning skills.
To address this, we carefully designed the options to achieve an appropriate balance.
For continuous features, we modeled the distribution of values using a Gaussian Mixture Model (GMM) with three components ($n = 3$).
Then, one component was selected by sampling according to the posterior probability of the true value given the GMM.
We then calculated a margin by multiplying the standard deviation of the selected component by a difficulty constant, which clinicians recommended setting to 2.
Using this margin, we constructed an arithmetic sequence centered around the correct answer to generate the full set of answer choices.
Post-processing was applied to eliminate implausible options, such as negative lab values, ensuring that all choices remain clinically realistic.
For multi-class features, answer choices were randomly sampled based on their frequency distribution.

\subsection{Training}
We trained LLMs using questions generated from real-world clinical records, employing the Group Relative Policy Optimization (GRPO) algorithm \citep{shao2024deepseekmath}, which was also used in training Deepseek-R1, a state-of-the-art general domain reasoning model \citep{guo2025deepseek}.
During preliminary experiments, we observed that when denoising masked values, the model's responses began directly with the answer, which made it difficult to follow the underlying reasoning process.
To encourage the model to reason before answering, we adopted a technique known as zero-shot Chain-of-Thought (CoT) prompting \citep{kojima2022large}.
By having them start their responses with the phrase "Let's think step by step," the model delayed making a final decision until it had worked through the reasoning process.
We retained most of the original GRPO hyperparameters, modifying only the number of generated reasoning traces per question (from 1024 to 7) and the batch size (from 1024 to 35) to accommodate our computational resource constraints.
These changes were found to be empirically stable.
The training was conducted on eight NVIDIA A100 80GB GPUs for approximately one week.
We performed full-parameter fine-tuning rather than using parameter-efficient methods.

\subsection{Performance Evaluation}
Extracting answers from natural language reasoning is challenging, so enforcing strict formatting is a common approach.
Although we prompted the models to follow a specific format (\textbackslash$boxed\{\}$), they occasionally failed to do so, particularly among similarly-sized baseline models.
Since our primary goal is to evaluate the model's clinical reasoning capabilities rather than its ability to adhere to formatting instructions, we assessed only the correctness of the answer, regardless of formatting.
Following Kojima et al. \citep{kojima2022large}, we appended the phrase ``Therefore, the answer is'' to the model's reasoning.
We then selected the multiple-choice option (A–E) with the highest log probability.
This strategy forces the model to choose one of the given options, and we compute evaluation metrics based on that selection.

Since our model was trained with zero-shot CoT, we also applied zero-shot CoT prompting during evaluation of other models to ensure fairness.
However, for reasoning-specialized models (o3-mini-high, Deepseek-R1, QwQ-32B, and Deepseek-R1-distilled-Qwen-14B), this step was unnecessary and thus omitted.

Due to the nature of the clinical dataset, some values are missing.
If the target feature to be masked is missing, we skipped the denoising.
As a result, denoising cannot be applied to test set patients for those features.
For example, the ``Appropriateness of Initial Empirical Therapy'' task has 6 missing values, while the ``ECOG at Discharge'' task has 110 missing values in the test set.
Therefore, the reported performance metrics are calculated based on 994 and 890 samples, respectively.

\begin{figure*}
    \includegraphics[width=1.0\linewidth]{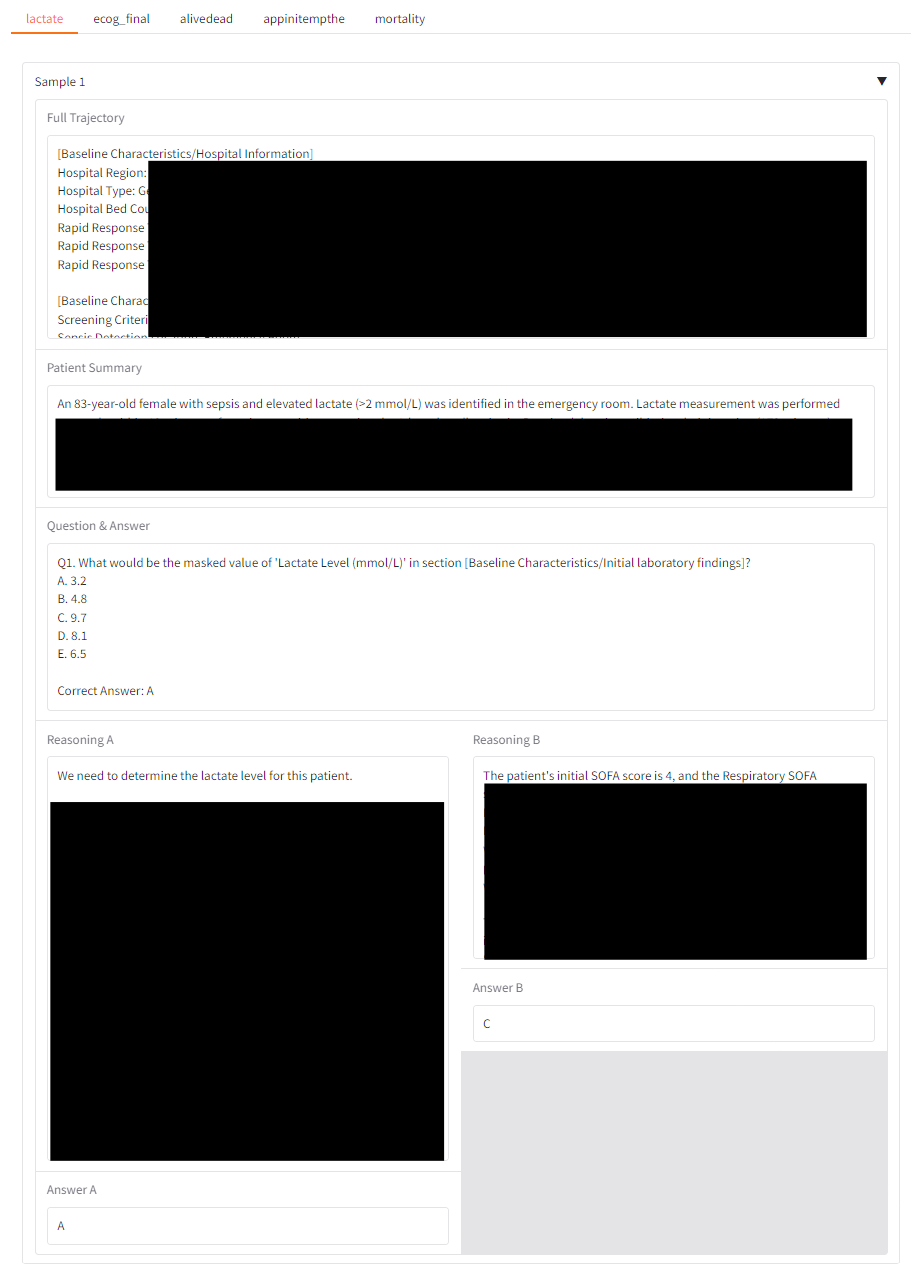}
    \caption{Sepsis Registry Expert Evaluation UI}
    \label{fig:ui}
\end{figure*}

\subsection{Expert Evaluation}

For the expert evaluation on the sepsis registry and MIMIC-III data, we provided the full patient trajectory, which was identical to what was given to the models, the ground truth answer, and the responses and choices from two models.
Additionally, we included a patient trajectory summary generated by GPT-4o \citep{achiam2023gpt} to assist annotators.
To ensure fairness, we randomly shuffled the responses from Phi-4 and \mname.
The expert evaluation interface is shown in Figure \ref{fig:ui}.
Evaluators were instructed to compare the two responses side by side and select the one whose clinical reasoning was more sound, judging both the final answer and the rationale behind it, and to additionally note any hallucinations or factually unreliable statements.
For the four sepsis registry tasks, in which each case was evaluated by all three intensivists, pairwise inter-rater agreement ranged from 60\% to 80\% per task, and Gwet's AC1 ranged from 0.33 to 0.75 across tasks.
For the antibiotic consultation task, each evaluator reviewed 20 non-overlapping cases, so inter-rater agreement could not be computed.

For the consultations on antibiotics use task, the original notes were written in Korean.
Since transferring this data outside the hospital is prohibited, we used the Llama-3.3-70B-Instruct model \citep{grattafiori2024llama} with FP8 quantization to perform the translation.
The same model was also used to segment the consultation responses into recommendation, assessment, and opinion sections.
Annotators were provided with the original recommendation, assessment, or opinion.

\section{Additional Base Model}\label{apd:base}

\begin{table}[!t]\centering
\caption{Results of applying our pipeline to an additional base model, Llama-3.1-8B. We report accuracy, with F1 score in parentheses for binary prediction tasks. Every evaluation except the sepsis registry denoising is zero-shot.}\label{tab:base_model}
\setlength{\tabcolsep}{0.4em}
\resizebox{\columnwidth}{!}{
\begin{tabular}{llrr}\toprule
Dataset &Task &Llama-3.1-8B &+ Ours \\\midrule
\multirow{4}{*}{Sepsis Registry} &Initial Lactate Den. &0.444 &\textbf{0.701} \\
&ECOG at Discharge Den. &0.423 &\textbf{0.547} \\
&Discharge Status Den. &\textbf{0.827} &0.810 \\
&App. Ini. Emp. Den. &0.764 &\textbf{0.895} \\\midrule
MIMIC-III &In-Hospital Mortality Pred. &0.179 (0.181) &\textbf{0.864 (0.218)} \\\midrule
Hospitalized Cohort &48h AKI Pred. &0.785 (0.424) &\textbf{0.810 (0.463)} \\\midrule
\multirow{2}{*}{Stroke Registry} &3-months mRS Pred. &0.575 &\textbf{0.628} \\
&1-year MACE Pred. &0.482 (0.213) &\textbf{0.642 (0.251)} \\\bottomrule
\end{tabular}
}
\end{table}

Our pipeline has no model-specific components and applies to any base model.
To verify that the observed gains are not specific to Phi-4, we additionally applied it to Llama-3.1-8B-Instruct \citep{grattafiori2024llama}, a smaller model from a different model family.
During initial training with GRPO, the reasoning of Llama-3.1-8B-Instruct collapsed to zero tokens, a known failure mode of GRPO that can also be attributed to the weaker reasoning of Llama-3.1-8B-Instruct relative to Phi-4.
We therefore replaced GRPO with DAPO \citep{yu2026dapo}, after which training finished without collapse.
As shown in Table \ref{tab:base_model}, the trained Llama-3.1-8B-Instruct improves over its base model on most tasks, including tasks where the F1 score decreased for Phi-4 (e.g., the 1-year MACE prediction).
These results indicate that our framework is robust to the choice of base model and reasoning algorithm, and provide further evidence of its cross-disease generalizability.

\newpage
\section{Data Statistics}\label{apd:stat}
\begin{table}[!htbp]\centering
\caption{Sepsis Registry Statistics. Values are presented as number (percentage) or median (interquartile range), as appropriate.}\label{tab:stat_sepsis}
\footnotesize
\begin{tabular}{lrrr}\toprule
\multicolumn{2}{c}{Feature} &Value \\\cmidrule{1-3}
\multirow{5}{*}{Age} &$<$50 &894 (7.5) \\\cmidrule{2-3}
&50-59 &1326 (11.1) \\\cmidrule{2-3}
&60-69 &2499 (20.9) \\\cmidrule{2-3}
&70-79 &3529 (29.5) \\\cmidrule{2-3}
&$\geq$80 &3733 (31.2) \\\cmidrule{1-3}
\multirow{2}{*}{Sex} &Male &6904 (57.6) \\\cmidrule{2-3}
&Female &5077 (42.4) \\\cmidrule{1-3}
\multicolumn{2}{c}{Septic Shock} &2163 (18.1) \\\cmidrule{1-3}
\multicolumn{2}{c}{SOFA Score} &6.0 (4.0–8.0) \\\cmidrule{1-3}
\multicolumn{2}{c}{Lactic Acid (mmol/L)} &2.6 (1.6–4.8) \\\cmidrule{1-3}
\multirow{6}{*}{Vital at Admission} &SBP (mmHg) &91.0 (80.0–111.0) \\\cmidrule{2-3}
&DBP (mmHg) &57.0 (48.0–68.0) \\\cmidrule{2-3}
&MBP (mmHg) &68.3 (58.7–83.3) \\\cmidrule{2-3}
&HR (rate/min) &106.0 (89.0–122.0) \\\cmidrule{2-3}
&RR (rate/min) &22.0 (20.0–26.0) \\\cmidrule{2-3}
&BT (°C) &37.2 (36.5–38.2) \\\cmidrule{1-3}
\multirow{7}{*}{Comorbidities} &Cardiovascular Disease &2752 (23.0) \\\cmidrule{2-3}
&Respiratory Disease &1721 (14.4) \\\cmidrule{2-3}
&Chronic neurologic Disease &3021 (25.2) \\\cmidrule{2-3}
&Chronic liver Disease &1105 (9.2) \\\cmidrule{2-3}
&Diabetes mellitus &4170 (34.8) \\\cmidrule{2-3}
&Chronic kidney Disease &1528 (12.8) \\\cmidrule{2-3}
&Connective tissue Disease &321 (2.7) \\\cmidrule{1-3}
\multirow{2}{*}{\makecell{Appropriateness of Initial\\Empirical Therapy}}&Appropriate &10516 (87.8) \\\cmidrule{2-3}
&Inappropriate &1364 (11.4) \\\cmidrule{1-3}
\multirow{5}{*}{Outcomes} &ICU Length of Stay (days) &4.0 (2.0–10.0) \\\cmidrule{2-3}
&ICU Mortality &1215 (10.1) \\\cmidrule{2-3}
&Hospital Length of Stay (days) &13.0 (7.0–25.0) \\\cmidrule{2-3}
&Hospital Mortality &3420 (28.5) \\\cmidrule{2-3}
&ECOG at Discharge &3.0 (2.0–5.0) \\\midrule
\bottomrule
\end{tabular}
\end{table}

\begin{table}[!htp]\centering
\caption{MIMIC-III Sepsis Cohort Statistics. Values are presented as number (percentage) or median (interquartile range), as appropriate.}\label{tab:stat_mimic}
\footnotesize
\begin{tabular}{lrrr}\toprule
\multicolumn{2}{c}{Feature} &Value \\\cmidrule{1-3}
\multirow{5}{*}{Age} &$<$50 &179 (17.9) \\\cmidrule{2-3}
&50-59 &171 (17.1) \\\cmidrule{2-3}
&60-69 &227 (22.7) \\\cmidrule{2-3}
&70-79 &200 (20.0) \\\cmidrule{2-3}
&$\geq$80 &223 (22.3) \\\cmidrule{1-3}
\multirow{2}{*}{Sex} &Male &527 (52.7) \\\cmidrule{2-3}
&Female &473 (47.3) \\\cmidrule{1-3}
\multirow{8}{*}{ICU Admission} &SOFA Score &4.0 (2.0–6.0) \\\cmidrule{2-3}
&Lactic Acid (mmol/L) &1.4 (1.0–1.9) \\\cmidrule{2-3}
&SBP (mmHg) &115.2 (105.6–127.2) \\\cmidrule{2-3}
&DBP (mmHg) &59.2 (52.6–66.3) \\\cmidrule{2-3}
&MBP (mmHg) &75.1 (68.7–82.7) \\\cmidrule{2-3}
&HR (rate/min) &84.4 (75.9–96.7) \\\cmidrule{2-3}
&RR (rate/min) &18.7 (16.5–21.7) \\\cmidrule{2-3}
&BT (°C) &36.8 (36.5–37.2) \\\cmidrule{1-3}
\multirow{6}{*}{Comorbidities} &Congestive Heart Failure &320 (32.0) \\\cmidrule{2-3}
&Chronic Pulmonary Disease &244 (24.4) \\\cmidrule{2-3}
&Renal Disease &185 (18.5) \\\cmidrule{2-3}
&Liver Disease &144 (14.4) \\\cmidrule{2-3}
&Diabetes &304 (30.4) \\\cmidrule{2-3}
&Cancer &73 (7.3) \\\cmidrule{1-3}
\multirow{4}{*}{Outcomes} &ICU Length of Stay (days) &2.9 (1.5–6.0) \\\cmidrule{2-3}
&ICU Mortality &56 (5.6) \\\cmidrule{2-3}
&Hospital Length of Stay (days) &8.4 (5.2–15.3) \\\cmidrule{2-3}
&Hospital Mortality &95 (9.5) \\\midrule
\bottomrule
\end{tabular}
\end{table}

\begin{table}[!htp]\centering
\caption{Hospitalized Cohort Statistics. Values are presented as number (percentage) or median (interquartile range), as appropriate.}\label{tab:stat_aki}
\footnotesize
\begin{tabular}{lrrr}\toprule
\multicolumn{2}{c}{Feature} &Value \\\cmidrule{1-3}
\multirow{5}{*}{Age} &$<$50 &174 (17.4) \\\cmidrule{2-3}
&50-59 &166 (16.6) \\\cmidrule{2-3}
&60-69 &226 (22.6) \\\cmidrule{2-3}
&70-79 &297 (29.7) \\\cmidrule{2-3}
&$\geq$80 &137 (13.7) \\\cmidrule{1-3}
\multirow{2}{*}{Sex} &Male &577 (57.7) \\\cmidrule{2-3}
&Female &423 (42.3) \\\cmidrule{1-3}
\multirow{6}{*}{Hospital Admission} &Baseline Creatinine (mg/dL) &0.9 (0.7–1.1) \\\cmidrule{2-3}
&Baseline eGFR (mL/min/1.73 m$^2$) &78.9 (60.9–96.9) \\\cmidrule{2-3}
&SBP (mmHg) &130.5 (111.8–147.6) \\\cmidrule{2-3}
&DBP (mmHg) &74.0 (65.3–83.0) \\\cmidrule{2-3}
&HR (rate/min) &83.0 (70.5–101.0) \\\cmidrule{2-3}
&BT (°C) &36.5 (36.2–36.9) \\\cmidrule{1-3}
\multirow{6}{*}{Comorbidities} &Congestive Heart Failure &41 (4.1) \\\cmidrule{2-3}
&Hypertension &132 (13.2) \\\cmidrule{2-3}
&Liver Disease &26 (2.6) \\\cmidrule{2-3}
&Diabetes &117 (11.7) \\\cmidrule{2-3}
&Renal Disease &33 (3.3) \\\cmidrule{2-3}
&Cancer &155 (15.5) \\\cmidrule{1-3}
\multirow{4}{*}{Outcomes} &AKI within 8 Days &191 (19.1) \\\cmidrule{2-3}
&Critical AKI within 8 Days &78 (7.8) \\\cmidrule{2-3}
&Hospital Length of Stay (days) &12.0 (5.0–90.0) \\\cmidrule{2-3}
&100 Days Mortality &76 (7.6) \\\midrule
\bottomrule
\end{tabular}
\end{table}

\begin{table}[!htp]\centering
\caption{Stroke Registry Statistics. Values are presented as number (percentage) or median (interquartile range), as appropriate.}\label{tab:stat_stroke}
\footnotesize
\begin{tabular}{lrrr}\toprule
\multicolumn{2}{c}{Feature} &Value \\\cmidrule{1-3}
\multirow{5}{*}{Age} &$<$50 &96 (9.6) \\\cmidrule{2-3}
&50-59 &181 (18.1) \\\cmidrule{2-3}
&60-69 &226 (22.6) \\\cmidrule{2-3}
&70-79 &286 (28.6) \\\cmidrule{2-3}
&$\geq$80 &211 (21.1) \\\cmidrule{1-3}
\multirow{2}{*}{Sex} &Male &589 (58.9) \\\cmidrule{2-3}
&Female &411 (41.1) \\\cmidrule{1-3}
\multirow{6}{*}{Hospital Admission} &NIHSS &3.0 (1.0–6.0) \\\cmidrule{2-3}
&mRS &0.0 (0.0–0.0) \\\cmidrule{2-3}
&SBP &147.0 (130.0–165.0) \\\cmidrule{2-3}
&Onset to Arrival Time (hrs) &12.2 (3.6–36.9) \\\cmidrule{2-3}
&IV Thrombolysis &103 (10.3) \\\cmidrule{2-3}
&Endovascular Treatment &86 (8.6) \\\cmidrule{1-3}
\multirow{6}{*}{Comorbidities} &Previous Stroke &204 (20.4) \\\cmidrule{2-3}
&Prevous Myocardial Infraction &1 (0.1) \\\cmidrule{2-3}
&Hypertension &653 (65.3) \\\cmidrule{2-3}
&Diabetes &327 (32.7) \\\cmidrule{2-3}
&Dyslipidemia &319 (31.9) \\\cmidrule{2-3}
&Atrial Fibrillation &176 (17.6) \\\cmidrule{1-3}
\multirow{6}{*}{Outcomes} &Hospital Length of Stays (days) &6.3 (4.4–10.0) \\\cmidrule{2-3}
&Hospital Mortality &8 (0.8) \\\cmidrule{2-3}
&NIHSS at Discharge &2.0 (0.0–5.0) \\\cmidrule{2-3}
&mRS at Discharge &2.0 (1.0–3.0) \\\cmidrule{2-3}
&3-Months mRS &1.0 (0.0–3.0) \\\cmidrule{2-3}
&1-Year MACE &107 (10.7) \\\midrule
\bottomrule
\end{tabular}
\end{table}

\section{Data Samples and Prompts}\label{apd:samples}
Here, we provide the sample and prompt used for each evaluation.
Since we permuted the values for de-identification, some of them may appear unrealistic.
Note that the samples are formatted specifically for Phi-4 and \mname.
For the other models, we used the appropriate formats accordingly.

\begin{tcolorbox}[breakable, title=Sepsis Registry Denoising Example]
    \small
    \begin{lstlisting}
<|im_start|>system<|im_sep|>Do not add a disclaimer or any other unnecessary sentences after the prediction.
Put your final answer (letter choice only) within \boxed{}.<|im_end|><|im_start|>user<|im_sep|>[Baseline Characteristics/Hospital Information]
Hospital Region: Non-capital Area
Hospital Type: Tertiary Hospital
Hospital Bed Count: 1001~1500
Rapid Response Team Activity: yes
Rapid Response Team Grade: Grade 1
Rapid Response Team Activity Time: 24 hours/day

[Baseline Characteristics/Screening Condition]
Screening Criteria: yes
Sepsis Detection Location: Emergency Room
Age Over 19: yes
qSOFA: yes
Respiratory Rate Over 22: yes
Systolic Blood Pressure Under 100: no
Altered Mental Status: Not measurable
Blood Culture Test: yes

[Baseline Characteristics/Eligibility Criteria]
Eligibility Criteria: yes
Sepsis: yes
Suspected or Confirmed Infection: yes
SOFA Score Over 2: yes
Septic Shock: no
Vasopressor Use: no
Lactate Over 2 mmol/L: no

[Baseline Characteristics/Basic Data of Study Participants]
Age: 85
Sex: Female
Height (cm): 152.0
Weight (kg): 37.9
Predicted Body Weight (kg): 48.1
BMI (kg/m^2): 16.65
ER Sepsis Recognition: no
Follow-up in Current Institution: no
Recent 90-day Hospitalization Over 2 Days: no
Nursing Home Residence: yes
Recent 30-day Antibiotic/Anticancer Treatment: no
Recent 30-day Wound Treatment: yes
Recent 30-day Dialysis Treatment: no
Comorbidity_Cardiovascular Disease: no
Comorbidity_Chronic Respiratory Disease: no
Comorbidity/Chronic Neurological Disease: yes
Comorbidity/Chronic Liver Disease: no
Comorbidity/Diabetes Mellitus: no
Comorbidity/Chronic Kidney Disease: yes
Comorbidity/Connective Tissue Disease: no
Comorbidity/Immunocompromised: no
Comorbidity/Hematologic Malignancy: no
Comorbidity/Solid Malignant Tumor: no
Charlson comorbidity index/Age:4 ≥80
Charlson comorbidity index/DM: No DM
Charlson comorbidity index/Liver Disease: no
Charlson comorbidity index/Solid Tumor: no
Charlson comorbidity index/AIDS: no
Charlson comorbidity index/Chronic Kidney Disease: yes
Charlson comorbidity index/Congestive Heart Failure: no
Charlson comorbidity index/Myocardial Infarction: no
Charlson comorbidity index/Chronic Obstructive Pulmonary Disease: no
Charlson comorbidity index/Peripheral Vascular Disease: no
Charlson comorbidity index/Cerebrovascular Disease: yes
Charlson comorbidity index/Dementia: no
Charlson comorbidity index/Hemiplegia: no
Charlson comorbidity index/Connective Tissue Disease: no
Charlson comorbidity index/Leukemia: no
Charlson comorbidity index/Lymphoma: no
Charlson comorbidity index/Peptic Ulcer Disease: no
Charlson Comorbidity Index Total: 5
Clinical Frailty Scale: 5
ECOG Performance Status: 1
Time Zero Datetime: 2005-05-12 11:28:00.000
Initial Vital Sign SBP (mmHg): 138.0
Initial Vital Sign DBP (mmHg): 90.0
Initial Vital Sign MBP (mmHg): 105.7
Initial Vital Sign Heart Rate (/min): 77.0
Initial Vital Sign Respiratory Rate (/min): 24.0
Initial Vital Sign Body Temperature (℃): 37.1
Initial SOFA Score: 2
Respiratory SOFA Subscore: 2.0
Coagulation SOFA Subscore: 0.0
Hepatic SOFA Subscore: 0.0
Cardiovascular SOFA Subscore: 0
Neurological SOFA Subscore: 0
Renal SOFA Subscore: 0.0

[Baseline Characteristics/Initial laboratory findings]
Lactate Level (mmol/L): 0.89
White Blood Cell Count (10^3/uL): 10.0
Neutrophil Percentage (%): 91.0
Absolute Neutrophil Count (/uL): 9200.0
Hemoglobin (g/dL): 13.1
Hematocrit (%): 32,5
Platelet Count (10^3/uL): 263.0
Sodium (mmol/L): 135.0
Potassium (mmol/L): 3.1
Chloride (mmol/L): 97.0
Blood Urea Nitrogen (mg/dL): 20.6
Creatinine (mg/dL): 0.32
Bilirubin (mg/dL): 0.69
AST (U/L): 17.0
ALT (U/L): 5.0
Albumin (g/dL): 3.5
Prothrombin Time (INR): 0.93
C-Reactive Protein (mg/dL): 0.85
Glucose (mg/dL): 90.0
Arterial pH: 7.47
PaCO2 (mmHg): 45.0
PaO2 (mmHg): 80.0
Bicarbonate (Arterial) (mmol/L): 29.6
Procalcitonin (ng/mL): 0.372
Troponin I or T (ng/mL): 0.024

[Baseline Characteristics/Echocardiography (within 24 hours from the time zero)]
Echocardiography: yes
LV Systolic Dysfunction: no

[Baseline Characteristics/Initial characteristics of infection]
Site of Infection (MOSAICS II): Pulmonary
Type of Infection (MOSAICS II): Nursing home acquired

[Baseline Characteristics/Surviving Sepsis Campaign bundles]
Lactate Level: yes
Lactate Level Datetime: 2005-05-12 13:35:00.000
Blood Culture: yes
Blood culture performed datetime: 2005-05-12 13:35:00.000
Antibiotics: yes
Antibiotics Administration Datetime: 2005-05-12 19:05:00.000
Bolus Fluid Infusion: no
Vasopressors: no
Follow up lactate level: no

[Baseline Characteristics/Characteristics of initial antibiotics treatment for sepsis]
Antibiotics use before sepsis diagnosis: yes
Initial Empirical Antibiotics after sepsis diagnosis: Beta-lactams, Carbapenem
Combination Antibiotics: yes

[Baseline Characteristics/Adjunctive corticosteroid treatment]
Corticosteroid Treatment: no
Corticosteroid Treatment Datetime: 2005-05-12 18:05:00
Corticosteroid Type: Fludrocortisone
Combination Corticosteroids: no

[Baseline Characteristics/Source control]
First Infection Source Control: no
First Non-Surgical Infection Source Control: no
Surgical Infection Source Control: yes

[Microbiology/Pathogen identification]
Pathogen Identification: no

[Microbiology/Pathogen(s) responsible for sepsis]
Pathogen Type: Bacteria
Gram Positive Bacteria Presence: yes
Gram Positive Bacteria Type: Non-S. aureus Staphylococcus spp.
Gram Negative Bacteria Presence: no
Atypical Bacteria Presence: no
Pathogen Count: 1.0
Bacteria Count: 2.0
Bacteria Specimen: Blood
Bacteria Test Method: Culture
MDR Pathogen: no
Patogen Description: Staphylococcus hominis, Staphylococcus capitis - Blood culture

[Microbiology/Appropriatness of initial empirical therapy]
Appropriateness of Initial Empirical Therapy: Appropriate

[SAPS3 at ICU Admission (SAPS3)/Box 1: Patient characteristics before ICU admission]
Age: 83.0
Cancer History: no
Cancer Treatment History: no
Hematologic Malignancy History: no
CHF History: no
Liver Cirrhosis History: no
AIDS History: no
Hospital Length of Stay Before ICU: ≥ 28
Location Before ICU: Emergency room
Vasoactive Drug Use Before ICU: no

[SAPS3 at ICU Admission (SAPS3)/Box 2]
ICU Admission Reason: Cardiovascular: All others (default)
ICU Admission Reason: Hepatic: All others (default)
ICU Admission Reason: Gastrointestinal: All others (default)
ICU Admission Reason: Neurologic: All others (default)
Planned ICU Admission: planned
Planned Surgery: Scheduled surgery
Surgery Site: Transplantation surgery ; Liver, Kidney, Pancreas, Kidney and pancreas, Transplantation other

[SAPS3 at ICU Admission (SAPS3)/Box 3]
Systolic Blood Pressure (mmHg): 150.0
Diastolic Blood Pressure (mmHg): 67.0
Heart Rate (/min): 90.0
Body Temperature (℃): 36.9
Respiratory Rate (/min): 11.0
GCS Score: 5.0
White Blood Cell Count (10^3/uL): 10.0
Platelet Count (10^3/uL): 223.0
Creatinine (mg/dL): 0.45
Bilirubin (mg/dL): 0.89
pH: 7.45
Mechanical Ventilation: no
Non-Mechanical Ventilation Patient: PaO₂ ≥ 60 mmHg
SAPS3 Total Score: 53.0
Predicted Mortality by SAPS3: 23.9

[ICU Day 1/ICU admisstion date/time]
ICU Admission Datetime: 2005-05-12 23:15:00.000

[ICU Day 1/Body temperature]
Body Temperature (℃): 36.9

[ICU Day 1/SOFA score]
Respiratory SOFA subscore: 2.0
Coagulation SOFA subscore: 0.0
Hepatic SOFA subscore: 1.0
Cardiovascular SOFA subscore: 0.0
Neurological SOFA subscore: 4.0
Renal SOFA subscore: 0.0
Total SOFA Score: 7.0

[ICU Day 1/Laboratory findings]
White Blood Cell Count (10^3/uL): 8.8
Neutrophil Percentage (%): 88.4
Absolute Neutrophil Count (/uL): 7771.2
Hemoglobin (g/dL): 10.3
Platelet Count (10^3/uL): 221.0
Creatinine (mg/dL): 0.27
Albumin (g/dL): 3.5
Total Bilirubin (mg/dL): 0.58
C-reactive Protein (mg/dL): 11.05
Arterial pH: 7.43
PaCO2 (mmHg): 53.0
PaO2 (mmHg): 146.0
FiO2: 0.3

[ICU Day 1/Resource used at ICU day 1]
Invasive Mechanical Ventilation Use: no
Noninvasive Ventilation Use: no
High-Flow Nasal Cannula Use: yes
Continuous Renal Replacement Therapy Use: no
Extracorporeal Membrane Oxygenation Use: no
Hemoperfusion Use: yes

[ICU Day 1/Medications]
Vasopressors Use: no
Norepinephrine Use: no
Epinephrine Use: no
Vasopressin Use: yes
Dopamine Use: no
Other Vasopressors Use: no
Inotropes Use: no
Dobutamine Use: no
Digoxin Use: yes
Milrinone Use: no
Analgesics Use: yes
Remifentanil Use: no
Fentanyl Use: no
Morphine Use: no
Other Analgesics Use: yes
Sedatives Use: no
Dexmedetomidine Use: no
Benzodiazepine Use: no
Propofol Use: no
Ketamine Use: no
Other Sedative Use: yes
Neuromuscular blocking agent Use: no
Cisatracurium Use: no
Vecuronium Use: yes
Rocuronium Use: no
Other Neuromuscular blocking agent Use: no

[ICU Day 1/Adjunctive corticosteroid]
Adjunctive Corticosteroid  Use: yes
Adjunctive Corticosteroid Type: Hydrocortisone
Adjunctive Corticosteroid Combination: yes

[ICU Day 1/Transfusions]
Transfusion: no

[ICU Day 1/Input and output]
Input before ICU admission (mL): 500.0
Ouput before ICU admission (mL): 660.0
Input (mL): 1837.0
Output (mL): 753.0

[ICU Day 2/ICU admisstion date/time]
...
[ICU outcomes/ICU discharge date/time]
ICU Discharge Datetime: 2005-06-10 16:55:00.000
ICU Length of Stay (Days): 3.0
ICU Discharge Survival Status: Alive
ICU Discharge Type: GW in same hospital

[ICU outcomes/Hemodynamic support at ICU discharge]
Hemodynamic Support at ICU Discharge: no

[ICU outcomes/Other interventions at ICU discharge]
Oxygen Support at ICU Discharge: no
Mechanical Ventilation at ICU Discharge: no
High-Flow Nasal Cannula at ICU Discharge: yes
Tracheostomy at ICU Discharge: no
Renal Replacement Therapy at ICU Discharge: yes

[ICU outcomes/Resource used during ICU stay]
Mechanical Ventilation: no
Noninvasive Ventilation: no
High-Flow Nasal Cannula: no
Continuous Renal Replacement Therapy: no
ECMO: no
Hemoperfusion: no

[ICU outcomes/Medical events during ICU stay]
Ventilator-Associated Pneumonia: no
Catheter-Related Bloodstream Infection: no
Catheter-Associated Urinary Tract Infection: yes
ARDS: no
Arrhythmia: yes
Bleeding Requiring Intervention: no
CPR: no

[Final Outcome/Medical events during ICU stay]
Hospital Admission Datetime: 2005-05-12 12:28:00.000
Hospital Discharge Datetime: 2020-06-10 10:45:00.000
Hospital Length of Stay (Days): 28.0
Transfer Details: Step-down referral
ECOG at Discharge: [MASK]
ICU Admission During Hospital Stay: yes
Life-Sustaining Treatment Suspension: no

[Derived variable/Variables Related to Sepsis Bundle Treatment]
1-Hour Bundle Success - Lactate Level: yes
1-Hour Bundle Success - Blood Culture: yes
1-Hour Bundle Success - Antibiotic Administration: no
1-Hour Bundle Success - Fluid Therapy: yes
1-Hour Bundle Success - Vasopressor Use: yes
3-Hour Bundle Success - Lactate Level: yes
3-Hour Bundle Success - Blood Culture: yes
Recent 3-Hour Bundle Success - Antibiotic Administration: no
Recent 3-Hour Bundle Success - Fluid Therapy: yes
Recent 3-Hour Bundle Success - Vasopressor Use: yes
Recent 6-Hour Bundle Success - Lactate Level: yes
Recent 6-Hour Bundle Success - Blood Culture: no
Recent 6-Hour Bundle Success - Antibiotic Administration: yes
Recent 6-Hour Bundle Success - Fluid Therapy: no
Recent 6-Hour Bundle Success - Vasopressor Use: yes
Recent 1-Hour Bundle Success: no
Recent 3-Hour Bundle Success: no
Recent 6-Hour Bundle Success: yes
Time to Antibiotic Administration (Minutes): 337.0
Time to Lactate Level Measurement (Minutes): 7.0
Time to Blood Culture (Minutes): 8.0
1-Hour Antibiotic Administration: no
1-Hour Lactate Level Measurement: yes
1-Hour Blood Culture: yes

[Derived variable/ICU-related Time Variables]
Time to ICU Admission (Minutes): 2043.0
1-Hour ICU Admission: no
3-Hour ICU Admission: no
6-Hour ICU Admission: no
ICU Length of Stay (Days): 3.0
Q1. What would be the masked value of 'ECOG at Discharge' in section [Final Outcome/Medical events during ICU stay]?
A. 4
B. 5
C. 0
D. 1
E. 3<|im_end|><|im_start|>assistant<|im_sep|>Let's think step by step.
\end{lstlisting}

\end{tcolorbox}

\begin{tcolorbox}[breakable, title=MIMIC-III Sepsis Cohort Value Prediction Example]
    \small
    \begin{lstlisting}
<|im_start|>system<|im_sep|>Do not add a disclaimer or any other unnecessary sentences after the prediction.
Put your final answer (letter choice only) within \boxed{}.<|im_end|><|im_start|>user<|im_sep|>[Patient Information]
Gender: Male
Age: 65.0
Readmission: No

[Time: Onset-8h]
Mechanical Ventilation: No
Maximum Vasopressor Dose over Recent 4h (mcg/kg/min of norepinephrine equivalent): 0
Weight (kg): 70.600
GCS: 15
HR (bpm): 71.589
Systolic BP (mmHg): 130.760
Mean BP (mmHg): 84.870
Diastolic BP (mmHg): 65.984
RR (breaths/min): 20.288
Temperature (°C): 36.500
FiO2: 0.400
Potassium (mEq/L): 5.800
Sodium (mEq/L): 137
Chloride (mEq/L): 99
Glucose (mg/dL): 119
Magnesium (mg/dL): 2.100
Calcium (mg/dL): 9.200
Hb (g/dL): 15.500
WBC Count (K/ul): 7.800
Platelet Count (K/ul): 233
PTT (sec): 34.100
PT (sec): 12.600
Arterial pH: 7.310
paO2 (mmHg): 182
paCO2 (mmHg): 59
Arterial BE (mEq/L): 1
HCO3 (mEq/L): 30
Arterial Lactate (mmol/L): 0.900
SOFA: 4
SIRS: 1
Shock Index: 0.546
PaO2/FiO2: 435.000
Cumulative Fluid Balance: 0
SpO2 (%): 88.333
BUN (mg/dL): 12
Creatinine (mg/dL): 0.700
SGOT (U/L): 15
SGPT (U/L): 8
Total Bilirubin (mg/dL): 1
INR: 1.200
Total Fluid Input (mL): 0
4-Hour Fluid Input (mL): 0
Total Fluid Output (mL): 0
4-Hour Fluid Output (mL): 0

[Time: Onset-4h]
...
Q. What will the Maximum Vasopressor Dose over Recent 4h (mcg/kg/min of norepinephrine equivalent) value likely be 24 hours after the last records?
A. 0.053
B. 0.23
C. 0
D. 0.628
E. 2.469<|im_end|><|im_start|>assistant<|im_sep|>Let's think step by step.


\end{lstlisting}
\end{tcolorbox}

\begin{tcolorbox}[breakable, title=MIMIC-III Sepsis Cohort In-Hospital Mortality Prediction Example]
    \small
    \begin{lstlisting}
<|im_start|>system<|im_sep|>Do not add a disclaimer or any other unnecessary sentences after the prediction.
Put your final answer (letter choice only) within \boxed{}.<|im_end|><|im_start|>user<|im_sep|>[Patient Information]
...
Q. Is the patient likely to die in the hospital?
A. Yes
B. No<|im_end|><|im_start|>assistant<|im_sep|>Let's think step by step.
\end{lstlisting}

\end{tcolorbox}

\begin{tcolorbox}[breakable, title=Hospitalized Cohort Denoising Example]
    \small
    \begin{lstlisting}
<|im_start|>system<|im_sep|>Do not add a disclaimer or any other unnecessary sentences after the prediction.
Put your final answer (letter choice only) within \boxed{}.<|im_end|><|im_start|>user<|im_sep|>[Baseline Characteristics]
Age: 75
Sex: Female
Body Mass Index: 35.12
ICU Admission: Yes
Baseline Creatinine: 0.7
Baseline eGFR: 68.19

[Underlying Disease]
Acute Myocardial Infarction: No
Congestive Heart Failure: No
Peripheral Vascular Disease: Yes
Cerebrovascular Disease: Yes
Dementia: No
Pulmonary Disease: Yes
Connective Tissue Disease: No
Peptic Ulcer Disease: No
Liver Disease: No
Severe Liver Disease: No
Diabetes: Yes
Diabetic Complication: No
Paraplegia: Yes
Renal Disease: No
Cancer: Yes
Metastatic Cancer: No
HIV Infection: No
Hypertension: Yes
Acute Kidney Injury: No
Charlson Comorbidity Index: 3

[Prescription History within 6 Months Before Admission]
Acyclovir: No
Aminoglycoside: No
Amphotericin: No
ARB: Yes
Beta-blocker: No
Calcium Channel Blocker: Yes
Cisplatin: Yes
Colistin: Yes
Cyclosporine: No
Diuretics: Yes
NSAIDs: No
Statins: Yes
Tacrolimus: No
Vancomycin: Yes
Vasopressor: No

[Day 1 00:00 - 08:00]
Albumin: 3.8
Bilirubin: 0.6
Blood Urea Nitrogen (BUN): 12.0
Calcium: 8.1
Chloride: 109.0
Creatine Kinase (CK): 88.0
Carbon Dioxide (CO2): 29.0
Creatinine: 0.7
C-Reactive Protein (CRP): 0.15
Glucose: 115.0
Aspartate Aminotransferase (GOT/AST): 21.0
Alanine Aminotransferase (GPT/ALT): 18.0
Hemoglobin: 13.8
Lipase: 8.0
Platelet Count: 265.0
Potassium: 3.1
Sodium: 143.0
Troponin: 1.0
White Blood Cell Count (WBC): 6.31
Systolic Blood Pressure (Max): 150.0
Diastolic Blood Pressure (Max): 68.0
Pulse Rate (Max): 108.0
Body Temperature (Max): 37.1
Systolic Blood Pressure (Avg): 150.0
Diastolic Blood Pressure (Avg): 63.5
Pulse Rate (Avg): 104.0
Body Temperature (Avg): 36.8
Systolic Blood Pressure (Min): 150.0
Diastolic Blood Pressure (Min): 59.0
Pulse Rate (Min): 100.0
Body Temperature (Min): 36.5
Acute Kidney Injury (AKI): No
Critical Acute Kidney Injury: No

[Day 1 08:00 - 16:00]
...
[Day 1]
ACE Inhibitor: No
Acyclovir: No
Aminoglycoside: No
Amphotericin: No
Angiotensin II Receptor Blocker: No
Beta-blocker: No
Calcium Channel Blocker: No
Cisplatin: Yes
Colistin: No
Cyclosporine: No
Diuretics: Yes
NSAIDs: No
Statins: Yes
Tacrolimus: No
Vancomycin: No
Vasopressor: Yes
Major Surgery: No
Minor Surgery: No
General Anesthesia: No
Non-general Anesthesia: No
Surgery Duration (minutes): 0.0
Dialysis: No
...
[Day 2 [00:00-08:00]
...
[Day 7]
...
[Final Outcomes]
Dialysis within 100 Days After Admission: No
Death within 100 Days After Admission: No
Exclusion - Death Date Error: No
ESRD Diagnosis within 100 Days After Admission: No
CAPD within 100 Days After Admission: No
AVF within 100 Days After Admission: No
Minimum Creatinine (1-3 Weeks After Admission): 0.75
Minimum Creatinine (1-5 Weeks After Admission): 0.75
Minimum Creatinine within 100 Days After Admission: 0.75
Q1. What would be the masked value of 'Carbon Dioxide (CO2)' in section [Day 4 16:00 - 24:00]?
A. 24
B. 39.3
C. 29.1
D. 34.2
E. 18.9<|im_end|><|im_start|>assistant<|im_sep|>Let's think step by step.
\end{lstlisting}
\end{tcolorbox}

\begin{tcolorbox}[breakable, title=AKI Prediction Example]
    \small
    \begin{lstlisting}
<|im_start|>system<|im_sep|>Do not add a disclaimer or any other unnecessary sentences after the prediction.
Put your final answer (letter choice only) within \boxed{}.<|im_end|><|im_start|>user<|im_sep|>[Baseline Characteristics]
...
Q. Would AKI occur in the next 48 hours?\nA.yes\nB.no<|im_end|><|im_start|>assistant<|im_sep|>Let's think step by step.
\end{lstlisting}
\end{tcolorbox}

\begin{tcolorbox}[breakable, title=Stroke Registry Denoising Example]
    \small
    \begin{lstlisting}
<|im_start|>system<|im_sep|>Do not add a disclaimer or any other unnecessary sentences after the prediction.<|im_end|><|im_start|>user<|im_sep|>Gender: Female
Age: 66
Onset date: 08/20/09 13:00:00
Time Last Known Well: 08/20/09 14:00:00
First abnormal time: 08/20/09 14:00:00
Time of Symptom Detection: 08/20/09 17:42:00
Index stroke: Ischemic Stroke
Height: 161.0
Weight: 54.0
BMI: 20.8
Initial NIHSS: 2
Previous mRS: 0.0
Arrival route: ER
Transfer-in: No
Onset situation: during sleep
Chief complaint: dysarthria
Stroke unit admission: yes
Education: 10-12 years
Ischemia or hemorrhage: hemorrhage
Ischemia TOAST classification: LAA
Hemorrhage IVH: No
Hemorrhage SAH: Yes
Hemorrhage SDH: No
TIA: No
Image positive: No
Risk Factor TIA: no
Risk Factor stroke: yes
Risk Factor type: Ischemic
Risk Factor PAD: no
Risk Factor CHD: no
Risk Factor HTN: no
Risk Factor DM: yes
Risk Factor DM details: diagnosed at admission
Risk Factor HL: yes
Risk Factor HL details: history of hl
Risk Factor smoking: no
Risk Factor AF: no
Potential Source of Cardioembolism (PSCE)/High Risk: no
Potential Source of Cardioembolism/High Risk/Mechanical prosthetic valve: no
Potential Source of Cardioembolism/High Risk/Mitral stenosis with atrial fibrillation: yes
Potential Source of Cardioembolism/High Risk/Atrial fibrillation (other than lone atrial fibrillation): no
Potential Source of Cardioembolism/High Risk/Left atrial/atrial appendage thrombus: no
Potential Source of Cardioembolism/High Risk/Sick sinus syndrome: no
Potential Source of Cardioembolism/High Risk/Recent myocardial infarction (<4 weeks): no
Potential Source of Cardioembolism/High Risk/Left ventricular thrombus: yes
Potential Source of Cardioembolism/High Risk/Dilated cardiomyopathy: no
Potential Source of Cardioembolism/High Risk/Akinetic left ventricular segment: no
Potential Source of Cardioembolism/High Risk/Atrial myxoma: no
Potential Source of Cardioembolism/High Risk/Infective endocarditis: yes
Potential Source of Cardioembolism/High Risk/Others: yes
Potential Source of Cardioembolism (PSCE)/Medium Risk: no
Potential Source of Cardioembolism/Medium Risk/Mitral valve prolapse: no
Potential Source of Cardioembolism/Medium Risk/Mitral annulus calcification: yes
Potential Source of Cardioembolism/Medium Risk/Mitral stenosis without atrial fibrillation: no
Potential Source of Cardioembolism/Medium Risk/Left atrial turbulence (smoke): no
Potential Source of Cardioembolism/Medium Risk/Atrial septal aneurysm: no
Potential Source of Cardioembolism/Medium Risk/Patent foramen ovale: no
Potential Source of Cardioembolism/Medium Risk/Atrial flutter: yes
Potential Source of Cardioembolism/Medium Risk/Lone atrial fibrillation: no
Potential Source of Cardioembolism/Medium Risk/Bioprosthetic cardiac valve: yes
Potential Source of Cardioembolism/Medium Risk/Nonbacterial thrombotic endocarditis: no
Potential Source of Cardioembolism/Medium Risk/Congestive heart failure: no
Potential Source of Cardioembolism/Medium Risk/Hypokinetic left ventricular segment: yes
Potential Source of Cardioembolism/Medium Risk/Myocardial infarction (>4 weeks, <6 months): no
History of medication - anti-platelet: yes
History of medication - clopidogrel: Yes
History of medication-anti coagulation: no
History of medication-hypertension: no
History of medication-anti hyperlipidemia-statin: no
History of medication-anti hyperlipidemia: no
History of medication-anti diabetes: no
First brain imaging time after arrival: 11/30/16 18:11:00
Brain imaging type-CT: Yes
Brain imaging type-MRI: Yes
Stroke Location/ICA: no
Stroke Location/MCA: no
Stroke Location/ACA: no
Stroke Location/PCA: no
Stroke Location/Basilar: no
Stroke Location/Vertebral: no
Stroke Location/SCA: no
Stroke Location/AICA: no
Stroke Location/PICA: no
Stroke Location/Multiple: No
Stroke Location/Negative: No
Stroke Location/Cortex: yes
Stroke Location/Cortex/Side: Lt
Stroke Location/Corona radiata: no
Stroke Location/BG or IC: no
Stroke Location/Thalamus: no
Stroke Location/Midbrain: no
Stroke Location/Pons: no
Stroke Location/Medulla: no
Stroke Location/Cerebellum: no
MR or CT Angiography/Stenosis at ACA: no
MR or CT Angiography/Stenosis at ACA/Status: No
MR or CT Angiography/Stenosis at MCA: [MASK]
MR or CT Angiography/Stenosis at MCA/Side: Lt
MR or CT Angiography/Stenosis at MCA/Status: No
MR or CT Angiography/Stenosis at PCA: no
MR or CT Angiography/Stenosis at PCA/Status: No
MR or CT Angiography/Stenosis at Basilar: no
MR or CT Angiography/Stenosis at Basilar/Status: No
MR or CT Angiography/Stenosis at Vertebral: no
MR or CT Angiography/Stenosis at Vertebral/Status: No
MR or CT Angiography/Stenosis at ExCrICA: no
MR or CT Angiography/Stenosis at ExCrICA/Status: No
MR or CT Angiography/Stenosis at InCrICA: no
MR or CT Angiography/Stenosis at InCrICA/Status: No
MR or CT Angiography/Stenosis at CCA: no
MR or CT Angiography/Stenosis at CCA/Status: No
MR or CT Angiography/Stenosis at Aortic arch: no
MR or CT Angiography/Stenosis at Aortic arch/Status: No
MR or CT Angiography/Multiple Stenosis: No
MR or CT Angiography/Negative Stenosis: No
Acute Endovascular Treatment: Not performed
IV tPA use: No
IV thrombolysis tPA dose: No
Endovascular Treatment drug - urokinase: No
Endovascular Treatment reoperation: No
Endovascular Treatment drug - tirofiban: No
Endovascular Treatment drug - other: No
Endovascular Treatment device - penumbra: No
Endovascular Treatment device - solitare: No
Endovascular Treatment device - merci: No
NIHSS 24 hours after thrombolysis: No
Vascular occlusion state: No
Vascular recanalization state: No
Acute Endovascular Treatment antiplt: yes
Acute Endovascular Treatment/aspirin: Yes
Acute Endovascular Treatment/clopidogrel: Yes
Acute Endovascular Treatment/aspirin + Dypiridamol: no
Acute Endovascular Treatment/cilostazol: no
Acute Endovascular Treatment/trifluzal: no
Acute Endovascular Treatment/ticlopidine: no
Acute Endovascular Treatment/others: no
Acute Endovascular Treatment anticoagulation: no
Acute Endovascular Treatment/Heparin: no
Acute Endovascular Treatment/warfarin: no
Treatment-acute Med (apixaban): no
Treatment-acute Med (dabigatran): no
Treatment-acute Med (rivaroxaban): no
Treatment-acute Med (edoxaban): no
Acute Endovascular Treatment/LMWH: no
Acute Endovascular Treatment/thrombin inhibitor: no
Acute Endovascular Treatment/others: no
Treatment-discharge med antiplt: yes
Treatment-Discharge Med (Aspirin): Yes
Treatment-Discharge Med (Clopidogrel): Yes
Treatment-discharge med (aspirin + Dypiridamol): no
Treatment-Discharge Med (Cilostazol): no
Treatment-Discharge Med (Triflusal): no
Treatment-Discharge Med (ticlopidine): no
Treatment-Discharge Med (others): no
Treatment-discharge med anticoagulation: no
Treatment-Discharge Med(warfarin): no
Treatment-Discharge Med(apixaban): no
Treatment-Discharge Med(dabigatran): no
Treatment-Discharge Med(rivaroxaban): no
Treatment-Discharge Med(edoxaban): no
Treatment-Discharge Med(LMWH): no
Treatment-Discharge Med(others): no
Treatment-intervention(decompressive surgery): no
Treatment-intervention(bypass surgery): no
Treatment-intervention(Endarterectomy): no
Treatment-intervention(Angioplasty): no
Treatment-intervention(Others): no
Medication for RF-Hyperlipidemia: Lipitor
Medication for RF-statin: yes
Medication for RF-others: Mucosta
CT: yes
CT Angio: yes
perfusion CT: yes
Studies-MRI: yes
Studies-MRA: yes
diffusion MRI: yes
Studies-Perf MRI: yes
Studies-TTE: yes
Studies-TEE: no
Studies-Holter: yes
Initial WBC Test Result: 4.65
Initial Total Cholesterol: 176.0
Initial BUN: 13.0
Initial Creatinine: 0.78
Initial Hemoglobin: 13.2
Initial Triglycerides: 125.0
Initial Hematocrit: 40.9
Initial HDL Cholesterol: 37.1
Initial Fasting Blood Sugar: 120.0
Initial Platelets: 318.0
Initial LDL Cholesterol: 117.0
Initial HbA1c: 6.6
Initial Prothrombin Time: 0.96
Initial CRP: 0.06
Initial Glucose: 121.0
ECG: normal
Initial Systolic BP: 130.0
Initial Diastolic BP: 71.0
Discharge Date: 2016-12-06
Discharge NIHSS: 3.0
Discharge mRS: 2.0
Discharge State: Discharge
Discharge-Sub: To Home
No END during admission: No
previous mRS: 2
Admission NIHSS: 6
END (Early Neurological Deterioration) 1 Existence: Yes
END1 Kind: Stroke progression
END1 Date: 12/01/16 05:50:00
NIHSS at END1: 2.0
END (Early Neurological Deterioration) 2 Existence: No
END (Early Neurological Deterioration) 3 Existence: No
3mo Contact Loss: No
3mo mRS: 3
3mo Date: 2009-08-20
3mo Drug Adherence: Yes
3mo Drug Adherence - No contact: No
3mo Informant: Family (text)
3mo Informant (text): Spouse
3mo Motivation - Had you ever forgotten to take your medication?: No
3mo Motivation - Have you ever failed to keep your medication on time?: Yes
3mo Motivation - Have you ever forgotten to pick up your prescribed medication on time?: Yes
3mo Knowledge - Have there been times when you didn't take your medication because you felt well?: No
3mo Knowledge - Have there been times when you didn't take your medication because you felt unwell?: No
3mo Knowledge - Are you aware of the long-term benefits of taking your medication as explained by your doctor?: Yes
3mo Amount of medication taken in the past month: 80.0
SBP within 1 to 6 months after onset: 142.0
DBP within 1 to 6 months after onset: 74.0
Date of BP examination: 08/20/09 00:00:00
TC within 1 to 6 months after onset: 159.0
TG within 1 to 6 months after onset: 93.0
HDL within 1 to 6 months after onset: 46.0
LDL within 1 to 6 months after onset: 93.0
Date of LDL Examination : 08/20/09 00:00:00
3mo No Clinical event: Yes
3mo Event 1 - Existence: No
3mo Event 2 - Existence: No
3mo Event 3 - Existence: No
1y Contact Loss: No
1y mRS: 2.0
1y date: 02/19/18 00:00:00
1y No clinical event: Yes
1y Event 1 - Existence: No
1y Event 2 - Existence: No
Initial NIHSS-Level of Consciousness: 0.0
Initial NIHSS-Response to Questions: 0.0
Initial NIHSS-Response to Commands: 0.0
Initial NIHSS-Best Gaze: 0.0
Initial NIHSS-Visual Field: 2.0
Initial NIHSS-Facial Palsy: 1.0
Initial NIHSS-Arm weakness (Rt): 0.0
Initial NIHSS-Arm weakness (Lt): 2.0
Initial NIHSS-Leg weakness Rt): 0.0
Initial NIHSS-Leg weakness (Lt): 1.0
Initial NIHSS-Limb Ataxia: 0.0
Initial NIHSS-Sensory Loss: 0.0
Initial NIHSS-Best Language: 0.0
Initial NIHSS-Dysarthria: 0.0
Initial NIHSS-Neglect: 0.0
Initial NIHSSSS-Total Score: 6.0
Initial NIHSS Subscore Existence: Some partial scores exist and others filled with 0
1y event - Major Adverse Cardiovascular Events: Yes
TOAST2 (ischemic stroke subtypes): small vessel occlusion
Q1. What would be the masked value of 'MR or CT Angiography/Stenosis at MCA'?
A. yes
B. no<|im_end|><|im_start|>assistant<|im_sep|>Let's think step by step.
\end{lstlisting}
\end{tcolorbox}

\begin{tcolorbox}[breakable, title=Stroke Registry 3-Months mRS Example]
    \small
    \begin{lstlisting}
<|im_start|>system<|im_sep|>Do not add a disclaimer or any other unnecessary sentences after the prediction.<|im_end|><|im_start|>user<|im_sep|>Gender: Female
...
Q. What will the mRS value be 3 months after admission?
A. 2
B. 0
C. 1
D. 6
\end{lstlisting}

\end{tcolorbox}

\begin{tcolorbox}[breakable, title=Stroke Registry 1-Year MACE Example]
    \small
    \begin{lstlisting}
<|im_start|>system<|im_sep|>Do not add a disclaimer or any other unnecessary sentences after the prediction.<|im_end|><|im_start|>user<|im_sep|>Gender: Female
...
Q. Would major adverse cardiovascular events (MACE) occur within one year after discharge?
A. Yes
B. No
\end{lstlisting}

\end{tcolorbox}

\section{Extended Results}\label{apd:res}

For the sepsis registry, there are a total of 600 features that have values for at least one patient.
We sorted the tasks by the accuracy of \mname and displayed the number of available samples in the test set.
Note that the numbers in the graph indicate the number of non-missing values in the test set.

\begin{figure}
    \centering
    \includegraphics[width=0.9\linewidth]{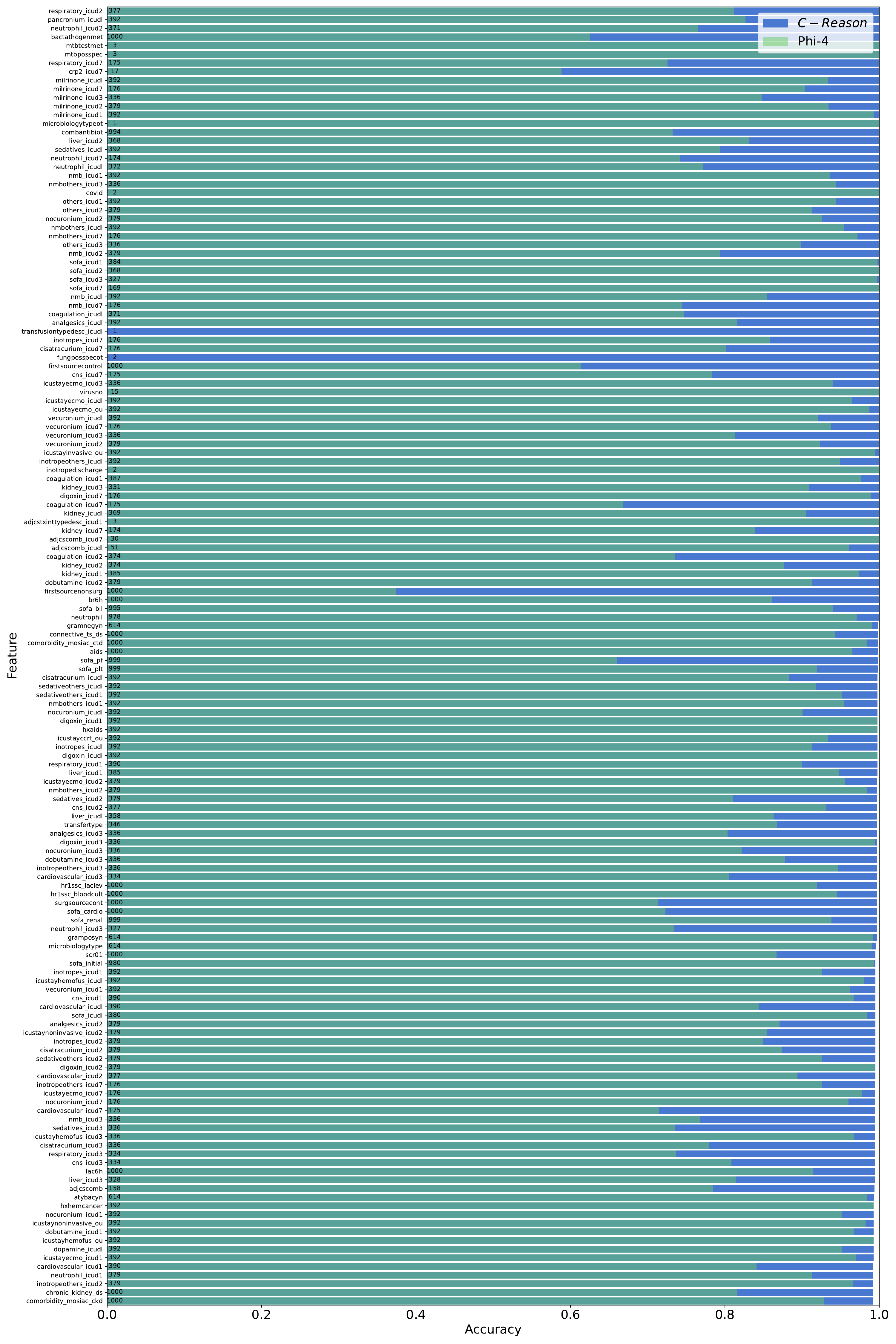}
    \caption{Sepsis Registry Per-Task Denoising Performance (1/4)}
    \label{fig:sup_1}
\end{figure}

\begin{figure}
    \centering
    \includegraphics[width=0.9\linewidth]{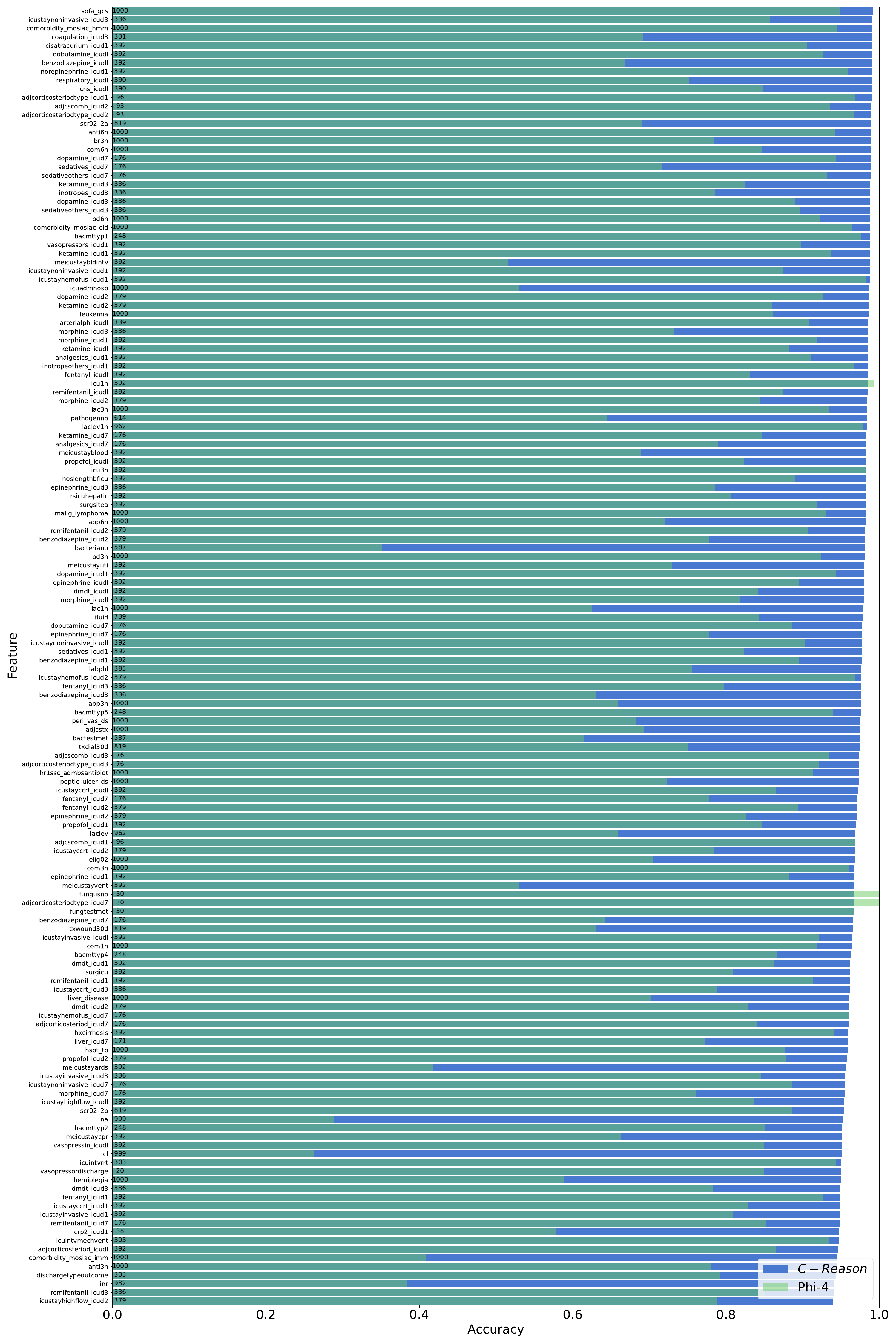}
    \caption{Sepsis Registry Per-Task Denoising Performance (2/4)}
    \label{fig:sup_2}
\end{figure}

\begin{figure}
    \centering
    \includegraphics[width=0.9\linewidth]{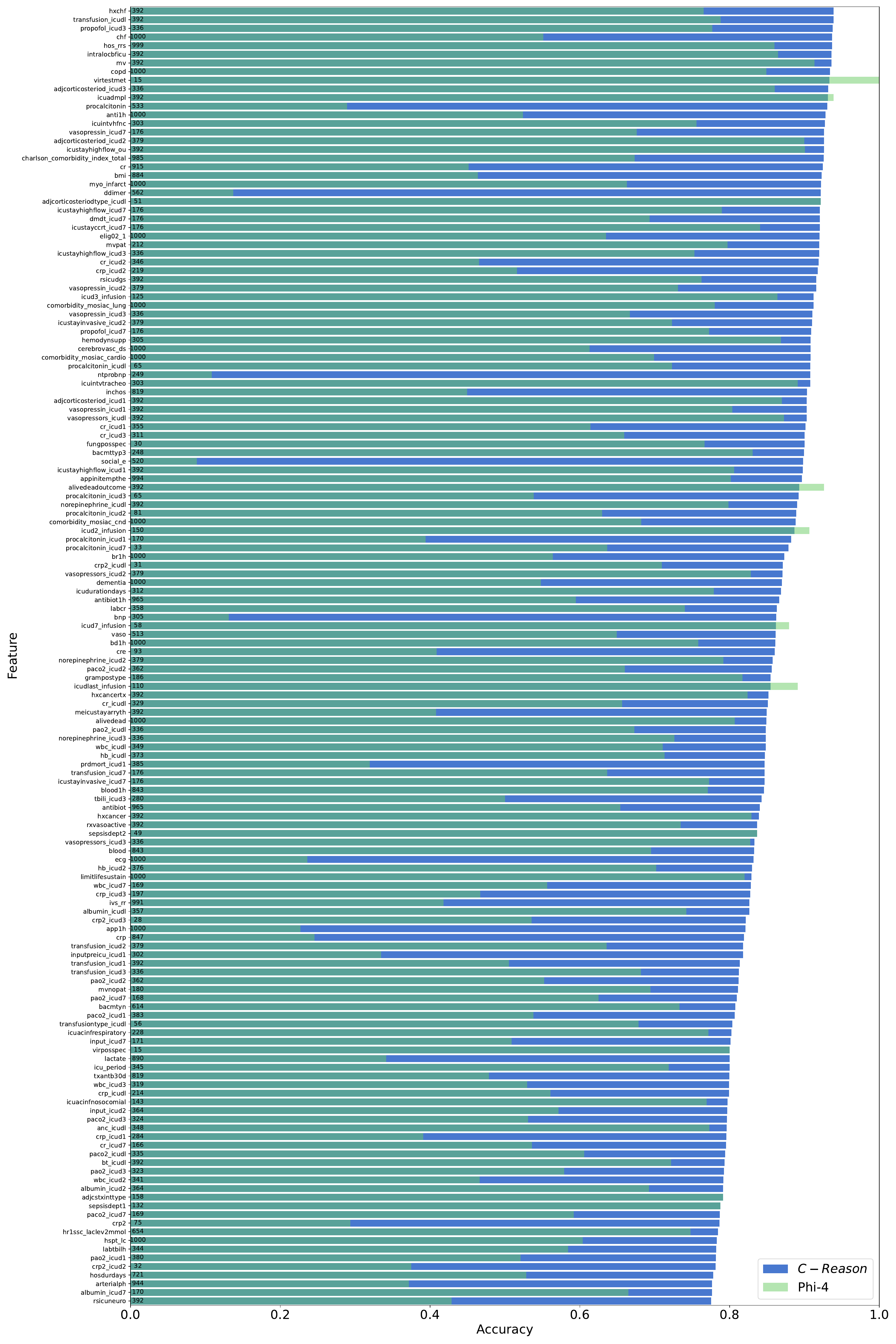}
    \caption{Sepsis Registry Per-Task Denoising Performance (3/4)}
    \label{fig:sup_3}
\end{figure}

\begin{figure}
    \centering
    \includegraphics[width=0.9\linewidth]{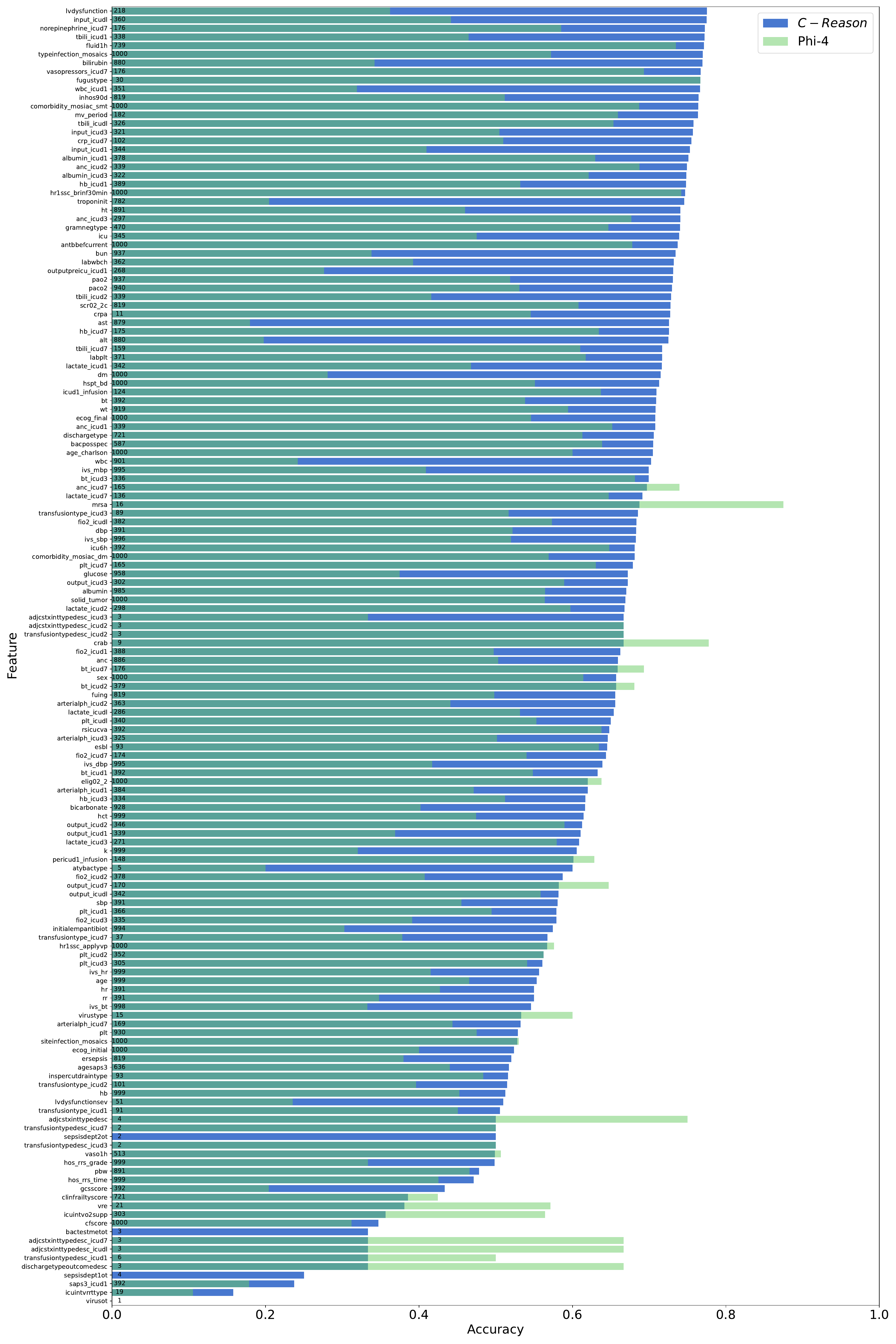}
    \caption{Sepsis Registry Per-Task Denoising Performance (4/4)}
    \label{fig:sup_4}
\end{figure}

\section{Failure Cases and Reliability}\label{apd:failure}

During the expert evaluation, we asked clinicians to flag any hallucinations or factually unreliable statements in addition to their preference.
We found no cases in which the model fabricated results that were absent from the input data.
However, we observed a fair number of cases in which the model misinterpreted causal relationships between features.
For example, a value of True for ``Life-Sustaining Treatment Suspension'' does not imply that the patient died, since the patient may subsequently have improved; nevertheless, the model often concluded that the patient had died.
Such misinterpretations, rather than outright fabrication, were the dominant failure mode.

\end{document}